# Internal Flow Signatures for Self-Checking and Refinement in LLMs

**Sungheon Jeong** [1]  **Sanggeon Yun** [1]  **Ryozo Masukawa** [1]  **Wenjun Haung** [1]  **Hanning Chen** [1]  **Mohsen Imani** [1]

## Abstract

Large language models can generate fluent answers that are unfaithful to the provided context, while many safeguards rely on external verification or a separate judge after generation. We introduce *internal flow signatures* that audit decision formation from depthwise dynamics at a fixed inter-block monitoring boundary. The method stabilizes token-wise motion via bias-centered monitoring, then summarizes trajectories in compact *moving* readout-aligned subspaces constructed from the top token and its close competitors within each depth window. Neighboring window frames are aligned by an orthogonal transport, yielding depth-comparable transported step lengths, turning angles, and subspace drift summaries that are invariant to within-window basis choices. A lightweight GRU validator trained on these signatures performs self-checking without modifying the base model. Beyond detection, the validator localizes a culprit depth event and enables a targeted refinement: the model rolls back to the culprit token and clamps an abnormal transported step at the identified block while preserving the orthogonal residual. The resulting pipeline provides actionable localization and low-overhead self-checking from internal decision dynamics. *Code is available on* GitHub.

## 1. Introduction

Large language models (LLMs) often produce fluent answers that are locally consistent yet globally incorrect (Huang et al., 2023; Maynez et al., 2020; Lin et al., 2022; Li et al., 2023). In practical deployments, a user typically needs to know not only whether an answer is wrong, but also whether the model was internally confident for the right reasons (Geng et al., 2024). Current safeguards lean on external verification, retrieval, or an additional LLM judge

[1]Department of Computer Science, University of California, Irvine. Correspondence to: Sungheon Jeong <sungheoj@uci.edu>.

(Nakano et al., 2021; Lewis et al., 2020; Zheng et al., 2023). These approaches add latency and cost, and they react only after the full generation is produced, while the internal process that formed the decision remains largely unobserved (Gao et al., 2023; Kossen et al., 2024; Farquhar et al., 2024).

We take a different view of inference. Rather than treating an autoregressive transformer as a single input-to-output mapping, we interpret generation as a depthwise flow of internal states and intermediate readouts (Belrose et al., 2023; Ferrando et al., 2023; Pal et al., 2023). A key lens is how logit competition evolves across depth (Belrose et al., 2023; Chuang et al., 2023): as a token decision forms, the top token and its close competitors shift in structured ways (Ferrando et al., 2023; Pal et al., 2023). This yields a measurable trace of decision formation that supports internal audits and lightweight self-checking (Elhage et al., 2021; Azaria & Mitchell, 2023; Chen et al., 2024a; Ji et al., 2024; Sriramanan et al., 2024). Unlike Logit lens (Elhage et al., 2021), which inspects what a layer encodes, we track how the decision trajectory is shaped across depth, enabling localization of atypical motion to specific internal updates.

Turning this intuition into reliable measurements faces two obstacles. First, any fixed global coordinate system can be misleading, since readout-relevant directions and what is linearly decodable can shift with depth (Belrose et al., 2023; Tenney et al., 2019; Ethayarajh, 2019). Second, the monitored boundary typically includes normalization with learned gain and bias, and its placement affects depthwise behavior, introducing depth dependent offsets that can contaminate token-wise motion and complicate depth aggregation (Ba et al., 2016; Xiong et al., 2020). Without resolving these issues, depthwise signatures become unstable and difficult to compare across layers, prompts, or models. This can prevent module level attribution of where the trajectory gets steered off course during generation.

We address these obstacles by stabilizing depthwise measurements in a local moving frame. We monitor the residual stream at a fixed boundary and remove depth dependent token shared offsets, so token motion reflects decision dynamics rather than layer specific shifts. Within each depth window, we build a compact readout aligned frame from the top token and close competitors, then align neighboring frames orthogonally so step size, turning, and drift remain comparable across depth.

These signatures enable self checking by training a lightweight validator on geometric flow patterns in depthwise decision formation. From flow features extracted during generation, the validator separates reliable and unreliable regimes by detecting atypical transported step length, turning, and subspace drift in a moving readout aligned frame, without modifying the base LLM. Across tasks and models, this supervision is stable and learnable, yielding consistent separation between non hallucination and hallucination like behaviors with minimal overhead. Beyond detection, the same signal localizes a single depth localized culprit event. We roll back to the token position where it occurs and regenerate while intervening at only one transformer block. The intervention clamps an abnormally large transported step in the readout aligned low dimensional frame while preserving the orthogonal residual component, producing targeted refinement that reduces hallucination without retraining or modifying the base model.

## 2. Related Work

**Hallucination detection and self-checking in LLMs.** Prior work mitigates hallucination and factual errors by adding external verification, including retrieval augmentation (Lewis et al., 2020; Nakano et al., 2021), post-hoc checking with an additional language model (Zheng et al., 2023; Li et al., 2023), and faithfulness benchmarks (Maynez et al., 2020; Lin et al., 2022). While effective, these approaches act only after generation, add latency and cost, and treat the base model as a black box, offering limited access to the internal decision process that produced the output.

**Logit-based probing and intermediate readouts.** Recent work probes transformer representations via intermediate readouts. Logit Lens (Elhage et al., 2021) and refinements such as Tuned Lens (Belrose et al., 2023) decode logits from hidden states across depth to study how token predictions evolve (Elhage et al., 2021; Chuang et al., 2023). Related analyses examine when semantic information becomes linearly decodable (Tenney et al., 2019; Ethayarajh, 2019) and how intermediate representations anticipate final outputs (Ferrando et al., 2023; Pal et al., 2023). While these methods offer layer-wise snapshots, we instead trace token-level decision formation as a depthwise trajectory, enabling localization of atypical motion to specific internal updates rather than static representational states.

**Representation geometry and depth-dependent subspaces.** Transformer representations exhibit anisotropy and depth-dependent geometric structure (Ethayarajh, 2019; Li et al., 2020), and task-relevant linear subspaces can shift substantially across layers, limiting a single global coordinate system (Tenney et al., 2019; Belrose et al., 2023). Recent analyses further suggest that representational change across depth encodes meaningful computation rather than noise (Pal et al., 2023). Motivated by these findings, we construct readout-aligned subspaces within depth windows and model their drift across depth, treating changes in the local frame as part of the signal rather than a nuisance.

**Internal signals for reliability and lightweight validation.** Recent work uses internal signals to assess reliability beyond external verification. Generation-time statistics and hidden-state dynamics can predict errors, overconfidence, or hallucination without modifying the base model (Kadavath et al., 2022; Manakul et al., 2023; Kuhn et al., 2023; Chen et al., 2024a), and related efforts leverage entropy, logit dynamics, or intermediate activations to assess validity or trigger selective abstention (Jiang et al., 2023; Chen et al., 2024b). While effective, these approaches often rely on task-specific heuristics or shallow statistics. We instead extract structured flow signatures that track how decisions evolve across depth and train a lightweight validator on these internal dynamics, without altering the base LLM or its decoding procedure.

**Inference-time refinement via internal interventions.** Recent work mitigates hallucination via inference-time interventions on model internals, including activation editing (Wang et al., 2025; Turner et al., 2023), layer-contrast decoding (Chuang et al., 2023), and hidden-state steering in vision-language models (Su et al., 2025; Wu et al., 2025; Liu et al., 2025). Our refinement uses flow signatures to localize a sample-specific culprit event in depth and token position, then intervenes only on that update while preserving the orthogonal component, coupling detection and refinement in a shared geometric frame.

## 3. Depthwise Flow Signatures

### 3.1. Setup, monitoring boundary, and objective

This subsection fixes the monitoring boundary and the quantities we record, yielding stable depth-comparable measurements for downstream validation.

**Model and monitored boundary state.** Consider an autoregressive transformer on a token sequence $x_{1:n}$ with $B$ blocks. For token position $t \in \{1, \ldots, n\}$ and boundary index $b \in \{0, \ldots, B\}$, let $h_{t,b} \in \mathbb{R}^d$ be the residual stream state observed at a fixed inter-block monitoring boundary, chosen once and kept fixed throughout the analysis. For each block $b \in \{0, \ldots, B-1\}$, let $o_{t,b} \in \mathbb{R}^d$ be the total attention contribution (summed over heads) and let $m_{t,b} \in \mathbb{R}^d$ be the MLP contribution. The monitored update across the boundary is

$$h_{t,b+1} = \mathcal{N}_{b+1}(h_{t,b} + o_{t,b} + m_{t,b}),$$

where $\mathcal{N}_{b+1}$ is the boundary normalization with its learned affine parameters. All subsequent measurements are derived from states observed at this fixed boundary.

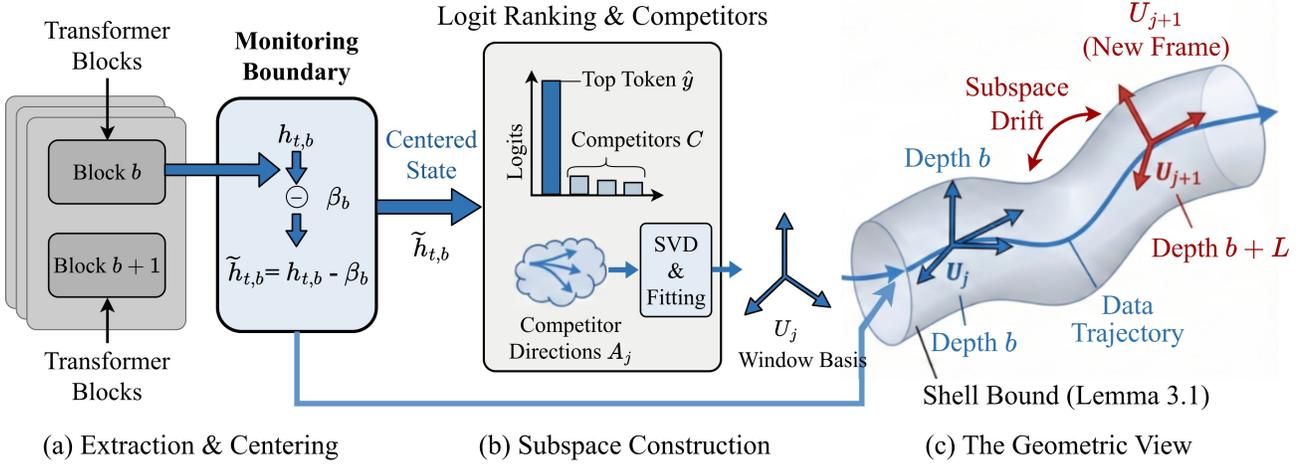

Figure 1. **Pipeline overview.** (a) **Extraction and centering.** We read states at a fixed monitoring boundary and apply bias centering. (b) **Subspace construction.** Logit ranking yields a top token and competitors, whose directions form a window basis via SVD. (c) **Geometric view.** The depthwise trajectory lies on a shell bound, while window frames evolve across depth, inducing subspace drift.

**Readout and competitor directions.** Let $W \in \mathbb{R}^{V \times d}$ be the readout to logits over a vocabulary of size $V$, and write $\ell_{t,b} = W h_{t,b} \in \mathbb{R}^V$. We use $W$ only to obtain depth-local competitor directions from the top ranked token and its nearest alternatives. These directions act as task-aligned probes in a shared parameter space, enabling depth-wise comparison without enforcing a single global subspace.

**Objective.** We monitor depthwise flow signatures without a single global readout subspace. Readout-relevant directions vary with depth, so we summarize token motion at the fixed boundary in window-specific readout-aligned frames and treat their drift as a signal for validation.

**Bias centered monitoring.** Let $\beta_b \in \mathbb{R}^d$ be the learned affine shift of the boundary normalization at depth $b$. We monitor the bias-centered state (Fig. 1a)

$$\tilde{h}_{t,b} = h_{t,b} - \beta_b. \tag{1}$$

For the depth step $b \to b+1$,

$$\tilde{h}_{t,b+1} - \tilde{h}_{t,b} = (h_{t,b+1} - h_{t,b}) - (\beta_{b+1} - \beta_b).$$

The offset $(\beta_{b+1} - \beta_b)$ depends on depth but not on token index, so it acts as a token-shared translation at each step. Centering removes this depth-dependent shift from token-wise motion and yields more stable pooled summaries across tokens. When needed, $\ell_{t,b} = W(\tilde{h}_{t,b} + \beta_b)$ with a depth-only offset $W\beta_b$.

**Lemma 3.1** (Shell bound from boundary normalization). *Assume the monitored boundary normalization admits the affine form*

$$\mathcal{N}_b(u_{t,b}) = \Gamma_b \mathcal{S}(u_{t,b}) + \beta_b, \quad \Gamma_b = \mathrm{diag}(\gamma_{b,1}, \dots, \gamma_{b,d}), \tag{2}$$

*with $0 < \gamma_{\min} \leq \gamma_{b,i} \leq \gamma_{\max}$ for all $b$ and $i$. Here $\mathcal{S}$ is the model's native normalization at the monitoring boundary (e.g., LayerNorm or RMSNorm). Assume further that on the monitored region the normalization denominator stays away from degeneracy so that*

$$c_{\min}\sqrt{d} \leq \|\mathcal{S}(u)\|_2 \leq c_{\max}\sqrt{d}, \tag{3}$$

*for constants $0 < c_{\min} \leq c_{\max}$. Then, for all $(t,b)$,*

$$\gamma_{\min} c_{\min} \sqrt{d} \leq \|\tilde{h}_{t,b}\|_2 \leq \gamma_{\max} c_{\max} \sqrt{d}. \tag{4}$$

*Proof.* Appendix A.2.

### 3.2. Competitor Directions and Moving Subspaces

Readout-relevant directions vary with depth, so a single global subspace yields unstable summaries. We instead fit windowed $k$-dimensional subspaces whose frames move with depth.

**Depth windows.** We index blocks by $b \in \{0, \dots, B-1\}$. Fix a window length $L \geq 1$ and a stride $s \geq 1$. For $j \geq 1$, define the window start $b_j = (j-1)s$ and the $j$th window

$$\mathcal{W}_j = \{b_j, \dots, b_j + L - 1\},$$

with the last window truncated so that it ends at $B-1$. Let $j(b)$ be the deterministic assignment that maps each block $b$ to the most recent window whose start satisfies $b_j \leq b$. This yields a unique $j(b)$ for every $b$. Exact endpoints and the closed-form $j(b)$ are given in Appendix A.3.

**Readout competitors within a window.** For each $(t,b)$, let $\widehat{y}_{t,b}$ be the top ranked token under the logit list $\ell_{t,b}$ and let $\mathcal{C}_{t,b}$ be the top $K$ competitors excluding $\widehat{y}_{t,b}$ (Appendix A.4).

For window $\mathcal{W}_j$, we collect competitor difference directions (Fig. 1**b**)

$$\mathcal{A}_j = \left\{ a_{t,b,y} = w_{\widehat{y}_{t,b}} - w_y \ : \ b \in \mathcal{W}_j, \ y \in \mathcal{C}_{t,b} \right\}. \quad (5)$$

We restrict $t$ to token positions permitted by a predefined mask $M$ (Appendix A.4).

**Subspace fitting.** From $\mathcal{A}_j$ in Eq. (5), we sample at most cap directions and stack them as rows into $D_j \in \mathbb{R}^{M_j \times d}$, where $M_j$ is the number of sampled directions. Let the compact SVD be $D_j = U_j^{(D)} \Sigma_j V_j^\top$ with right singular vectors $V_j = [v_{j,1}, \ldots, v_{j,d}]$. We take the window basis as the top $k$ right singular vectors,

$$U_j = [v_{j,1}, \ldots, v_{j,k}] \in \mathbb{R}^{d \times k}, \quad U_j^\top U_j = I_k, \quad (6)$$

and re-orthonormalize the extracted columns in finite precision. If $M_j = 0$, we use a deterministic fallback basis. Details of constructing $D_j$ (normalization, capping, and sampling), the re-orthonormalization step, and the deterministic handling of degenerate windows are given in Appendix A.5.

**Projected coordinates.** Using the assigned window basis $U_{j(b)}$ (Eq. 6) and the bias-centered state $\tilde{h}_{t,b}$ (Eq. 1), define the moving coordinates

$$p_{t,b} = U_{j(b)}^\top \tilde{h}_{t,b} \in \mathbb{R}^k. \quad (7)$$

These coordinates feed the transported step, turning, and drift signatures in the next subsection (Fig. 1**c**).

### 3.3. Transported Flow Signatures

#### 3.3.1. TRANSPORTED MOTION IN MOVING SUBSPACES

We align adjacent window frames with an orthogonal transport and record transported step length, turning, and a centered increment in $\mathbb{R}^k$.

**Window transport.** Window bases are only defined up to a within-window orthogonal rotation, so window switches can introduce spurious frame changes in $\mathbb{R}^k$. We align consecutive frames with the closest orthogonal map (Fig. 2**a**) (Fernando et al., 2013):

$$U_{j+1}^\top U_j = P_j \Sigma_j Q_j^\top, \quad R_{j \to j+1} = P_j Q_j^\top \in \mathbb{R}^{k \times k}.$$

Using the window assignment $j(b)$ from Sec. 3.2, define the step-wise transport

$$R_b = \begin{cases} I_k, & j(b+1) = j(b), \\ R_{j(b) \to j(b+1)}, & j(b+1) \neq j(b). \end{cases}$$

**Transported step and turning.** Using moving coordinates $p_{t,b}$ (Eq. 7), record the transported increment $\Delta p_{t,b} = p_{t,b+1} - R_b p_{t,b} \in \mathbb{R}^k$, and its step size

$$s_{t,b} = \|\Delta p_{t,b}\|_2. \quad (8)$$

For directional summaries, use $u_{t,b} = \frac{p_{t,b}}{\|p_{t,b}\|_2 + \varepsilon}$ with $\varepsilon_{\text{num}} > 0$ (Appendix B.5), and record the transported turning angle

$$\theta_{t,b} = \angle \big( u_{t,b+1}, \ R_b u_{t,b} \big). \quad (9)$$

**Centered increment for aggregation.** For a fixed sample and depth step $b$, remove a token-shared shift by centering $\{\Delta p_{t,b}\}_t$. Let $\mu_b \in \mathbb{R}^k$ be a rotation-equivariant robust center over masked token positions (Appendix A.7), and record

$$\Delta p_{t,b}^c = \Delta p_{t,b} - \mu_b, \quad s_{t,b}^c = \|\Delta p_{t,b}^c\|_2. \quad (10)$$

#### 3.3.2. COMPONENT CONTRIBUTIONS UNDER BOUNDARY NORMALIZATION

We record pre-normalization component magnitudes and post-normalization effective updates in the same target frame. Because boundary normalization is nonlinear, we use a path-integrated update $\Delta q_{t,b}$ to explain the transported increment $\Delta p_{t,b}$, and record the residual mismatch $\eta_{t,b}$.

**Target frame and pre-normalization magnitudes.** For step $b \to b+1$, use the next-frame basis $U_b^{\text{tgt}} = U_{j(b+1)} \in \mathbb{R}^{d \times k}$, so component projections share the same reference as the transported step at depth $b+1$. Project pre-normalization components: $\Delta p_{t,b}^o = (U_b^{\text{tgt}})^\top o_{t,b}$; $\Delta p_{t,b}^m = (U_b^{\text{tgt}})^\top m_{t,b}$. Record their magnitudes

$$a_{t,b} = \|\Delta p_{t,b}^o\|_2, \quad m_{t,b}^{\text{mag}} = \|\Delta p_{t,b}^m\|_2. \quad (11)$$

These quantify injected energy along the target subspace, prior to boundary-normalization reweighting.

**Path-integrated effective updates in the target frame.** Let $h_{t,b}^{\text{raw}}$ be the uncentered pre-normalization state and $\text{inj}_{t,b} = o_{t,b} + m_{t,b}$. Along the injection path $x_{t,b}(\alpha) = h_{t,b}^{\text{raw}} + \alpha \, \text{inj}_{t,b}$ for $\alpha \in [0,1]$, path integration captures state dependent scaling through the boundary map (Sundararajan et al., 2017).

Let $J_{b+1}(\cdot)$ denote the Jacobian of the boundary normalization at depth $b+1$. Record path-integrated channel updates

$$\Delta h_{t,b}^{\text{attn}} = \int_0^1 J_{b+1}\big(x_{t,b}(\alpha)\big) \, o_{t,b} \, d\alpha,$$

$$\Delta h_{t,b}^{\text{mlp}} = \int_0^1 J_{b+1}\big(x_{t,b}(\alpha)\big) \, m_{t,b} \, d\alpha,$$

evaluate $J_{b+1}(u)v$ via JVPs (Pearlmutter, 1994), and project to the target frame

$$\Delta q_{t,b}^{\text{attn}} = (U_b^{\text{tgt}})^\top \Delta h_{t,b}^{\text{attn}}, \quad \Delta q_{t,b}^{\text{mlp}} = (U_b^{\text{tgt}})^\top \Delta h_{t,b}^{\text{mlp}}.$$

Combine

$$\Delta q_{t,b} = \Delta q_{t,b}^{\text{attn}} + \Delta q_{t,b}^{\text{mlp}}, \quad c_{t,b} = \|\Delta q_{t,b}\|_2. \quad (12)$$

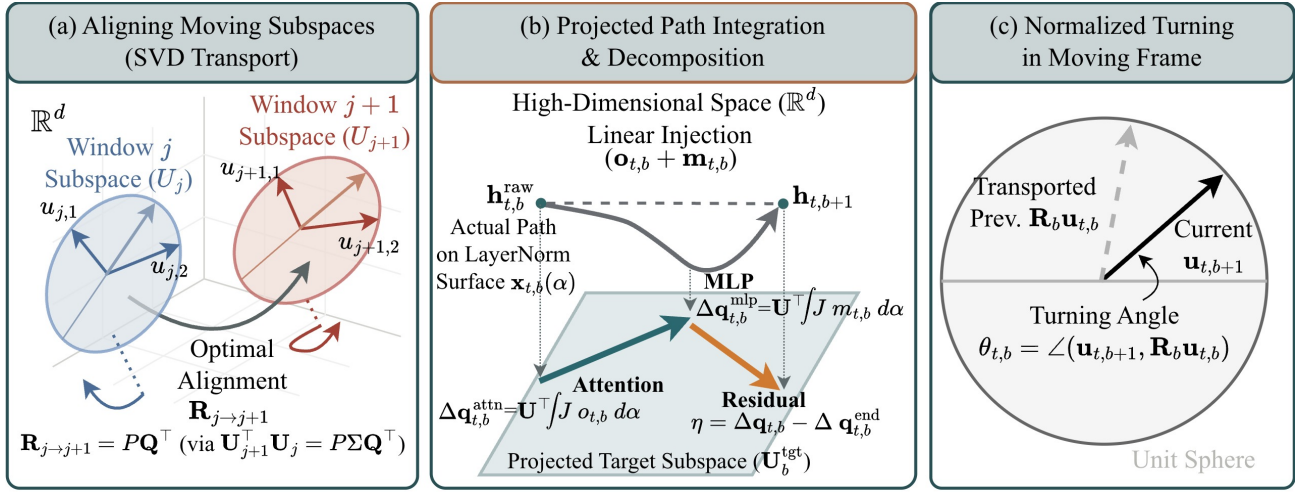

*Figure 2.* **Flow signatures from transported subspace trajectories. (a) Aligning moving subspaces.** Adjacent window bases $U_j$ and $U_{j+1}$ are aligned by an orthogonal transport $R_{j\to j+1}$ from the SVD of $U_{j+1}^\top U_j$, enabling consistent coordinates across depth windows. **(b) Projected path decomposition.** Boundary-normalized updates are projected onto a target window frame, producing attention and MLP contributions ($\Delta q_{t,b}^{\text{attn}}, \Delta q_{t,b}^{\text{mlp}}$) and a residual term $\eta_{t,b}$. **(c) Normalized turning in a moving frame.** The turning angle $\theta_{t,b}$ compares the current direction to the transported previous direction on the unit sphere, yielding a frame-invariant curvature summary.

**Single-point linearization and residual ratio.** A single-point JVP at $\alpha=1$ applied to the full injection gives

$$\Delta h_{t,b}^{\text{end}} = J_{b+1}(x_{t,b}(1))\,\text{inj}_{t,b}, \quad \Delta q_{t,b}^{\text{end}} = (U_b^{\text{tgt}})^\top \Delta h_{t,b}^{\text{end}}.$$

Record the deviation in $k$ space:

$$\eta_{t,b} = \Delta q_{t,b} - \Delta q_{t,b}^{\text{end}} \in \mathbb{R}^k, \; \rho_{t,b} = \frac{\|\eta_{t,b}\|_2}{\|\Delta q_{t,b}\|_2 + \varepsilon}. \quad (13)$$

In all experiments, $\Delta q_{t,b}$ uses a three-node Simpson rule over $\alpha \in \{0, 0.5, 1\}$, while $\Delta q_{t,b}^{\text{end}}$ uses a single JVP at $\alpha = 1$ (Appendix B.7).

**Component turning ratios.** Let $u_{t,b}^{\text{tgt}} = u_{t,b+1}$. We attribute bending to attention versus MLP by comparing their perpendicular components relative to $u_{t,b}^{\text{tgt}}$. For any $v \in \mathbb{R}^k$, remove its component along $u_{t,b}^{\text{tgt}}$: $v^\perp = v - \langle u_{t,b}^{\text{tgt}}, v\rangle u_{t,b}^{\text{tgt}}$. Record per-step ratios

$$r_{t,b}^{\text{attn}} = \frac{\|(\Delta q_{t,b}^{\text{attn}})^\perp\|_2}{\|(\Delta q_{t,b})^\perp\|_2 + \varepsilon}, \quad r_{t,b}^{\text{mlp}} = \frac{\|(\Delta q_{t,b}^{\text{mlp}})^\perp\|_2}{\|(\Delta q_{t,b})^\perp\|_2 + \varepsilon}.$$

We aggregate these per-step ratios into token-level summaries using a robust masked aggregation over depth (Appendix B.3), yielding stable summaries under outlier steps and variable token lengths (Fig.2c).

### 3.3.3. SUBSPACE DRIFT AND GAUGE INVARIANCE

**Subspace drift across windows.** Window bases are only identifiable up to within-window orthogonal rotations, so we measure drift with projector-level quantities that depend on the subspace itself (Gong et al., 2012). We record the Grassmann drift, $d_G(U_j, U_{j+1}) = \|U_j U_j^\top - U_{j+1} U_{j+1}^\top\|_2$, where $\|\cdot\|_2$ is the spectral norm (Appendix A.8).

We also record a state-coupled drift using a single deterministic anchor block $b_j^\star$ per window (we fix $b_j^\star$ to the window end), which keeps the cost linear in the number of windows while coupling drift to visited directions:

$$\chi_{t,j} = \frac{\|(U_{j+1} U_{j+1}^\top - U_j U_j^\top)\,\tilde{h}_{t,b_j^\star}\|_2}{\|\tilde{h}_{t,b_j^\star}\|_2 + \varepsilon}, \quad D_t = \sum_{j=1}^{J-1} \chi_{t,j}. \quad (14)$$

Appendix A.8 also gives the deterministic anchor rule and simple bounds such as $\chi_{t,j} \leq d_G(U_j, U_{j+1})$.

**Lemma 3.2** (Gauge invariance of transported signatures). *Fix window bases $\{U_j\}_j$ with orthonormal columns. For any orthogonal $\{Q_j\}_j$, set $U_j' = U_j Q_j$ and apply the same constructions to obtain the transported quantities under $\{U_j'\}$. Then following are invariant under $\{U_j\} \mapsto \{U_j'\}$:*

- *transported step lengths $\|\Delta p_{t,b}\|_2$,*
- *transported turning angles $\angle(u_{t,b+1}, R_b u_{t,b})$,*
- *any ratio built from Euclidean norms after removing the component along $u_{t,b}^{\text{tgt}}$ (including the perpendicular channel ratios and their masked depth aggregates),*
- *any drift depending only on projectors $U_j U_j^\top$ (including $d_G$ and the anchor-coupled drift),*
- *the residual ratio $\rho_{t,b}$ in Eq. 13.*

*If $\Delta p_{t,b}^c$ uses a rotation-equivariant center in $\mathbb{R}^k$ at each depth, then $\|\Delta p_{t,b}^c\|_2$ is also invariant. Rotation-equivariant* *choices include the Euclidean mean and the geometric median; coordinate-wise medians are not.*

**Proof.** Fix a window-wise basis change $U'_j = U_j Q_j$ with orthogonal $Q_j \in \mathbb{R}^{k \times k}$. For adjacent windows, write the compact SVD

$$U_{j+1}^\top U_j = P\Sigma Q^\top, \quad R_{j \to j+1} = PQ^\top.$$

Then
$$(U'_{j+1})^\top U'_j = Q_{j+1}^\top (U_{j+1}^\top U_j) Q_j$$

admits a compact SVD with $P' = Q_{j+1}^\top P$ and $Q' = Q_j^\top Q$, which yields

$$R'_{j \to j+1} = P'(Q')^\top = Q_{j+1}^\top R_{j \to j+1} Q_j.$$

Therefore the moving coordinates rotate as

$$p'_{t,b} = (U'_{j(b)})^\top \tilde{h}_{t,b} = Q_{j(b)}^\top p_{t,b}, \quad \Delta p'_{t,b} = Q_{j(b+1)}^\top \Delta p_{t,b}.$$

Since $Q_{j(b+1)}$ is orthogonal, it preserves Euclidean norms and angles. This proves invariance of the transported step length $\|\Delta p_{t,b}\|_2$ and the transported turning angle $\angle(u_{t,b+1}, R_b u_{t,b})$. Perpendicular ratios are invariant because the target direction and all target-frame projections undergo the same orthogonal rotation, preserving inner products and perpendicular norms. Projector-based drifts are invariant since $U'_j(U'_j)^\top = U_j U_j^\top$. $\rho_{t,b}$ is invariant because $\Delta q_{t,b}$ and $\Delta q_{t,b}^{\text{end}}$ share the same target-frame rotation. □

### 3.4. Self-checking via Flow Signature Validation

We train a lightweight validator $g_\phi$ that performs self-checking from the flow signatures in Sec. 3.3. For each sample $n$, the extractor produces a masked depth-and-token sequence $\{x_{n,b,t}\}$ together with a validity mask $M_n$. The validator maps the sequence to a single score $\hat{y}_n = g_\phi(\{x_{n,b,t}\}, M_n) \in [0, 1]$, used for threshold-based flagging at inference. Supervision is binary, $y_n \in \{0, 1\}$, obtained from dataset labels or an external judging protocol applied to the final answer, and $g_\phi$ is trained with a standard binary classification loss.

**Signature inputs.** Each event vector $x_{n,b,t}$ concatenates the scalar summaries $(s_{t,b}, s_{t,b}^c, \theta_{t,b})$ from Eqs. 8, 9, 10 $(a_{t,b}, m_{t,b}^{\text{mag}}, c_{t,b}, \rho_{t,b})$ from Eqs. 11, 12, 13, and the token-level drift summary $D_t$ from Eq. 14. All inputs are invariant to within-window basis rotations (Lemma 3.2), so $\hat{y}_n$ is comparable across depths and samples. Validator architecture, masking and packing, optimization, and calibration are deferred to the Appendix B.

### 3.5. Flow-Guided Refinement

Flow signature validation not only flags a sample but also localizes a culprit event along depth. Let the validator emit per-event scores $\{z_j\}$ with a validity indicator $\{I_j\}$. We select $j^\star \in \arg\max_{j: I_j=1} z_j$, $(t_0, b_0) = (t_{j^\star}, b_{j^\star})$, roll back the generation to the prefix $x_{1:t_0}$, and regenerate the continuation while intervening only at depth $b_0$; Fig. 3a. We keep the overall decode budget fixed: after rolling back to $x_{1:t_0}$, we regenerate only the remaining suffix budget, i.e., at most $T - t_0$ new tokens, so the final continuation length matches the original generation length.

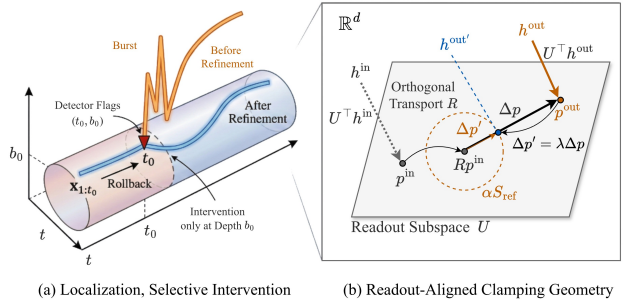

(a) Localization, Selective Intervention  (b) Readout-Aligned Clamping Geometry

*Figure 3.* **Flow-guided single-block refinement.** (a) Localizing a culprit event $(t_0, b_0)$, then an intervention at depth $b_0$. (b) Clamp an abnormally large transported step in a readout-aligned subspace.

The refinement targets the same depth-localized burst pattern that drives detection, an unusually large transported step inside a readout-aligned coordinate system; Fig. 3b. At depth $b_0$, let $h^{\text{in}}, h^{\text{out}} \in \mathbb{R}^d$ denote the transformer block input and output at the monitored boundary for the current decoding step. Here $U$ is a local readout-aligned basis fitted at the culprit decoding step from the current top token and its $K$ competitors, and it is then kept fixed during suffix regeneration (Appendix A.9). Using an orthonormal basis $U \in \mathbb{R}^{d \times k}$ together with an orthogonal transport $R \in \mathbb{R}^{k \times k}$ (Appendix A.9), we form

$$p^{\text{in}} = U^\top h^{\text{in}}, \quad p^{\text{out}} = U^\top h^{\text{out}}, \quad \Delta p = p^{\text{out}} - Rp^{\text{in}}.$$

A calibration pass on the rolled back prefix provides a reference step scale $s_{\text{ref}}$ (Appendix B.11). We apply an upper clamp with ratio $\alpha > 1$: $\lambda = \min\left(1, \frac{\alpha s_{\text{ref}}}{\|\Delta p\|_2 + \varepsilon}\right)$, $\Delta p' = \lambda \Delta p$, $p^{\text{out}'} = Rp^{\text{in}} + \Delta p'$. Finally, we rewrite only the $U$ component of the block output while keeping the orthogonal residual unchanged:

$$h^{\text{out}'} = h^{\text{out}} + U(p^{\text{out}'} - p^{\text{out}}).$$

All other depths remain untouched (Appendix A.9, B.11).

## 4. Experiments

We train a lightweight GRU-based validator to self-check from internal flow signatures, predicting whether a generated answer is hallucinated relative to the provided context. We evaluate on HaluEval (Li et al., 2023) across four tasks (QA, Dialogue, Summarization, General) and five base LLM families (Gemma2 (Team et al., 2024), Phi-3 (Abdin et al., 2024), LLaMA3 (Grattafiori et al., 2024), Qwen2.5 (Hui et al., 2024), Mistral (Jiang et al., 2024); each (task, model) split uses an 8:2 train:test prompt split.

Table 1. Hallucination detection performance on HaluEval across LLMs. We report classification Accuracy (%) and AUROC (%).

| Task | Accuracy | | | | | AUROC | | | | |
|---|---|---|---|---|---|---|---|---|---|---|
| | Qwen2.5 | Gemma 2 | Phi-3 | LLaMA3 | Mistral | Qwen2.5 | Gemma 2 | Phi-3 | LLaMA3 | Mistral |
| QA | 72.68 | 70.05 | 67.48 | 64.00 | 61.40 | 76.45 | 66.70 | 67.57 | 65.65 | 61.67 |
| General | 68.02 | 64.31 | 68.33 | 71.41 | 69.50 | 69.80 | 65.17 | 65.79 | 67.07 | 67.01 |
| Summarization | 53.81 | 54.48 | 53.82 | 52.12 | 57.27 | 58.17 | 63.32 | 59.08 | 59.05 | 58.85 |
| Dialogue | 50.75 | 56.02 | 56.88 | 47.84 | 54.83 | 51.83 | 54.13 | 58.12 | 54.68 | 54.23 |

Table 2. Hallucination ratios on HaluEval under refinement applied to hallucination-labeled samples only. Each cell reports Initial / Refine with relative reduction (%).

| Task | Qwen 2.5 | Gemma 2 | Phi-3 | LLaMA3 | Mistral |
|---|---|---|---|---|---|
| QA | 15.25 / 10.95 (↓ 28.19%) | 14.90 / 12.34 (↓ 17.18%) | 12.65 / 6.55 (↓ 48.22%) | 15.10 / 7.70 (↓ 48.99%) | 14.05 / 11.70 (↓ 16.73%) |
| General | 11.96 / 8.75 (↓ 26.86%) | 6.76 / 5.76 (↓ 14.79%) | 7.64 / 6.31 (↓ 17.41%) | 13.07 / 11.74 (↓ 10.17%) | 10.85 / 10.19 (↓ 6.08%) |
| Summarization | 40.90 / 40.20 (↓ 1.71%) | 39.80 / 37.10 (↓ 6.78%) | 42.15 / 40.70 (↓ 3.44%) | 45.80 / 44.60 (↓ 2.62%) | 49.70 / 47.25 (↓ 4.93%) |
| Dialogue | 50.95 / 48.10 (↓ 5.59%) | 39.95 / 34.95 (↓ 12.52%) | 51.90 / 46.30 (↓ 10.79%) | 43.60 / 34.30 (↓ 21.33%) | 44.50 / 39.05 (↓ 12.25%) |

For each prompt, we generate one answer, replay the prompt plus realized continuation through the same extractor with generated tokens fed back as fixed inputs (Bengio et al., 2015), and convert boundary traces into a masked flow-event sequence with depth as the time axis (Appendix B.8). Binary labels in 0, 1 come from ChatGPT(OpenAI, 2025).

The validator outputs a score $\hat{y} \in [0, 1]$ via mask-aware pooling over depth. We report Accuracy (threshold 0.5) and AUROC, and handle class imbalance by positive-class weighting (Details in Appendix B).

### 4.1. Hallucination Detection

#### 4.1.1. DETECTING RESULTS

Table 1 evaluates whether internal flow signatures alone provide a usable signal for hallucination detection. We observe consistent separability on **QA** and **General** across model families, with AUROC typically above 0.65, indicating that hallucination often co-occurs with structured depthwise flow changes that a lightweight validator can learn.

The task gap is informative. QA and General are settings where correctness is frequently determined by context-grounded verification, so hallucinated generations more often exhibit repeatable depth-localized deviations in transported motion, turning, and readout-aligned drift. In contrast, **Summarization** and **Dialogue** are intrinsically harder for the base LLMs, where errors tend to appear as subtle local factual edits or borderline multi-turn inconsistencies; such cases weaken alignment between the external labels and a stable internal flow signature, reducing separability. Overall, these results do not claim universal detection across tasks, but establish that internal flow provides a practical signal for self-checking in contexts where hallucination manifests as a concentrated trajectory deviation.

#### 4.1.2. HALLUCINATION REPORTING

In **QA**, hallucination often manifests as a depth-localized burst, where the transported increment $\|\Delta p_{t,b}\|_2$ spikes within a narrow depth band. This yields larger step lengths $s_{t,b}$ and higher turning $\theta_{t,b}$, with the effective update magnitude $c_{t,b}$ increasing in the same neighborhood; drift summaries $D_t$ also become more concentrated around the burst (Fig. 4a). In **General**, we observe a localized regime change where $s_{t,b}$ and $\theta_{t,b}$ rise together with $c_{t,b}$ and projected magnitudes such as $a_{t,b}$ and $m_{t,b}^{\mathrm{mag}}$, often with a more concentrated $D_t$. In both tasks, separability weakens when hallucination stays in a late diffuse or accumulation regime, where $(s_{t,b}, \theta_{t,b})$ and $D_t$ spread over deeper bands and overlap with non hallucination (Fig. 4b).

In **Summarization**, hallucination differs mainly by whether motion concentrates into a depth-gated band: $\Delta p_{t,b}$ becomes focused within a narrow depth neighborhood with increased $s_{t,b}$, $\theta_{t,b}$, and $c_{t,b}$ and a more localized $D_t$, while non hallucination more often remains diffuse and low-energy across later depths (Fig. 4c). When hallucination remains diffuse rather than concentrating, separability weakens. In **Dialogue**, depth-structured motion provides a more stable handle: we observe both a high-mass burst regime with elevated $s_{t,b}$, $\theta_{t,b}$, $c_{t,b}$, and concentrated $D_t$, and a low-mass diffuse regime whose profiles overlap with non hallucination. A practical summary is the burst depth band and the mixture between burst and diffuse modes, since the diffuse mode is often geometrically similar to non hallucination (Fig. 4d; Appendix C).

### 4.2. Flow-Guided Refinement

Table 2 reports hallucination ratios before and after flow-guided refinement, evaluated on samples labeled as hallucination. Refinement reduces hallucination rates most strongly on QA and more moderately on General, with rel-ative reductions reaching 48.99% on QA and 26.86% on General in the best cases.

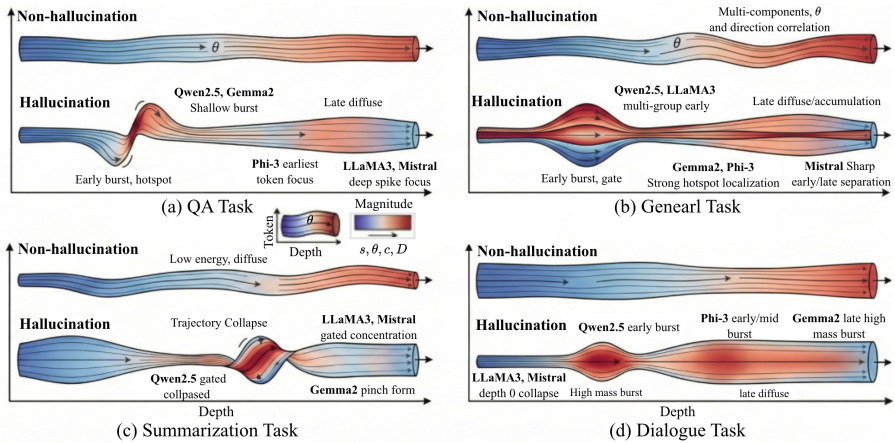

*Figure 4.* **Depthwise flow signatures across tasks.** Panels contrast non hallucination (top) and hallucination (bottom) by summarizing transported trajectories over depth and tokens across tasks. Color encodes event magnitude and curvature, and annotations mark regimes such as early depth bursts, late diffuse accumulation, and trajectory collapse.

The effect is task dependent. Summarization changes remain small across models (1.71%–6.78%), while Dialogue shows intermediate reductions (5.59%–21.33%). This trend matches the locality of the intervention: targeting a single depth is most helpful when the error is driven by a localized depth event, while longer-form behaviors that distribute deviations across tokens and depths admit less headroom from a single-block correction. We additionally report regeneration-only and random-depth variants in Appendix D.2 to separate pure regeneration effects from depth-specific intervention benefits.

### 4.3. Runtime Cost and Scalability

Refinement keeps the final continuation length fixed at $T$ tokens, but may discard a short suffix and regenerate it after an intervention point. Let $t_{\text{cur}}$ be the number of tokens already generated when the intervention triggers and let $t_0 \leq t_{\text{cur}}$ be the prefix length we keep fixed. With a cached decoding state at $t_0$, the extra decoding cost is exactly the discarded suffix length $t_{\text{cur}} - t_0$, so the total decoded tokens become $T + (t_{\text{cur}} - t_0)$ and the overhead factor is $1 + \frac{t_{\text{cur}} - t_0}{T}$, which is bounded by $2\times$ in the worst case. If cached restore at $t_0$ is unavailable, the method additionally replays the prefix to reconstruct state, which can add up to another $t_0$ tokens of forward passes per intervention.

## 5. Discussion

**Performance depends on base model competence.** The signatures are extracted from the same base LLM, so separability is bounded by the strength and consistency of its internal decision dynamics. When the base model handles the task reliably (QA and General in our setting), hallucinations more often align with consistent depth-local regime changes in the tracked signals, yielding stronger separation. For tasks the base model finds harder (Dialogue and Summarization), trajectories are more diffuse and label is ambiguous, weakens alignment between external labels and internal signatures and degrades validator performance.

**Refinement is a first step, not a final mechanism.** Our refinement uses a deliberately simple operator: a single-block clamp that shrinks an abnormally large transported step at a validator-localized culprit event while preserving the orthogonal residual. The heterogeneous gains across tasks reflect this locality: when the error is driven by a depth-localized deviation, a targeted correction can improve the trajectory with minimal collateral damage, whereas long-form settings such as Summarization often involve deviations distributed across tokens and depths, leaving less headroom for a single-site intervention. These results position refinement as an initial control primitive; extending it to multiple sites or short staged corrections can be the next direction, and it can naturally guide the choice of intervention locations.

## 6. Conclusion

We presented a method that audits how a language model forms an answer by monitoring its internal behavior during generation. Using these internal signals, a small validator can predict whether an output is likely to be hallucinated without changing the base model, and it can also indicate where the failure emerges inside the model. Building on this localization, we introduce a lightweight refinement step that intervenes at a single point to reduce hallucinations with minimal additional computation. Overall, internal monitoring offers a practical handle for detection and correction, and motivates stronger, more general interventions.

## Acknowledgments

This work was supported in part by the DARPA Young Faculty Award, the National Science Foundation (NSF) under Grants #2127780, #2319198, #2321840, #2312517, and #2235472, #2431561, the Semiconductor Research Corporation (SRC), the Office of Naval Research through the Young Investigator Program Award, and Grants #N00014-21-1-2225 and #N00014-22-1-2067, Army Research Office Grant #W911NF2410360. Additionally, support was provided by the Air Force Office of Scientific Research under Award #FA9550-22-1-0253, along with generous gifts from Xilinx and Cisco.

## Impact Statement

This paper studies depthwise internal flow signatures to detect and characterize hallucination behavior in large language models. The primary intended impact is to improve reliability, auditing, and safety of model outputs in downstream applications by enabling earlier detection and correction of unsupported generations. Potential risks include misuse of internal monitoring signals to optimize deceptive generations, to tune jailbreak style behaviors, or to support surveillance like deployment practices that reduce user autonomy. We mitigate these risks by focusing on model agnostic measurements rather than attack recipes, by reporting limitations and failure cases, and by encouraging responsible disclosure and evaluation under safety guidelines when releasing artifacts.

# A. Technical Appendix

## A.1. Normalization band constants and sufficient conditions

This appendix pins down explicit constants in the norm band $c_{\min}\sqrt{d} \le \|\mathcal{S}(u)\|_2 \le c_{\max}\sqrt{d}$ used in Lemma 3.1, under simple second moment conditions compatible with common LayerNorm or RMSNorm implementations.

**Per token normalization notation.** For a vector $u \in \mathbb{R}^d$, fix $\varepsilon > 0$.

**LayerNorm case.** Write the empirical mean and centered second moment

$$\mu(u) = \frac{1}{d}\sum_{i=1}^{d} u_i, \qquad s_2(u) = \frac{1}{d}\sum_{i=1}^{d}\left(u_i - \mu(u)\right)^2.$$

Use

$$\mathcal{S}_{\mathrm{LN}}(u) = \frac{u - \mu(u)\mathbf{1}}{\sqrt{s_2(u) + \varepsilon}}.$$

**RMSNorm case.** Write the (uncentered) second moment

$$m_2(u) = \frac{1}{d}\sum_{i=1}^{d} u_i^2.$$

Use

$$\mathcal{S}_{\mathrm{RMS}}(u) = \frac{u}{\sqrt{m_2(u) + \varepsilon}}.$$

**Lemma A.1** (Exact norm identity and variance-to-band reduction (LayerNorm))**.** *For any $u \in \mathbb{R}^d$,*

$$\|\mathcal{S}_{\mathrm{LN}}(u)\|_2^2 = \frac{d\,s_2(u)}{s_2(u) + \varepsilon}. \tag{15}$$

*Consequently, if a monitored region satisfies a two-sided variance condition*

$$0 < v_{\min} \le s_2(u) \le v_{\max}, \tag{16}$$

*then $\|\mathcal{S}_{\mathrm{LN}}(u)\|_2$ lies in a $\sqrt{d}$-scaled band*

$$c_{\min}\sqrt{d} \le \|\mathcal{S}_{\mathrm{LN}}(u)\|_2 \le c_{\max}\sqrt{d}, \qquad c_{\min} = \sqrt{\frac{v_{\min}}{v_{\min} + \varepsilon}}, \quad c_{\max} = \sqrt{\frac{v_{\max}}{v_{\max} + \varepsilon}} \le 1. \tag{17}$$

**Lemma A.2** (Exact norm identity and moment-to-band reduction (RMSNorm))**.** *For any $u \in \mathbb{R}^d$,*

$$\|\mathcal{S}_{\mathrm{RMS}}(u)\|_2^2 = \frac{d\,m_2(u)}{m_2(u) + \varepsilon}. \tag{18}$$

*Consequently, if a monitored region satisfies*

$$0 < v_{\min} \le m_2(u) \le v_{\max}, \tag{19}$$

*then $\|\mathcal{S}_{\mathrm{RMS}}(u)\|_2$ lies in a $\sqrt{d}$-scaled band*

$$c_{\min}\sqrt{d} \le \|\mathcal{S}_{\mathrm{RMS}}(u)\|_2 \le c_{\max}\sqrt{d}, \qquad c_{\min} = \sqrt{\frac{v_{\min}}{v_{\min} + \varepsilon}}, \quad c_{\max} = \sqrt{\frac{v_{\max}}{v_{\max} + \varepsilon}} \le 1. \tag{20}$$

**Proof.** For LayerNorm,

$$\|\mathcal{S}_{\mathrm{LN}}(u)\|_2^2 = \sum_{i=1}^{d} \frac{(u_i - \mu(u))^2}{s_2(u) + \varepsilon} = \frac{d\, s_2(u)}{s_2(u) + \varepsilon},$$

which gives Eq. 15. Under Eq. 16, $x \mapsto x/(x + \varepsilon)$ is increasing on $(0, \infty)$, yielding Eq. 17. For RMSNorm,

$$\|\mathcal{S}_{\mathrm{RMS}}(u)\|_2^2 = \sum_{i=1}^{d} \frac{u_i^2}{m_2(u) + \varepsilon} = \frac{d\, m_2(u)}{m_2(u) + \varepsilon},$$

which gives Eq. 18; Eq. 20 follows from Eq. 19 by the same monotonicity argument.

**Sufficient conditions supplied by common implementations.** Eq. 17 follows on any monitored region where the empirical variance satisfies Eq. 16. Some implementations or custom kernels further enforce this by applying a variance floor or clamping:

- **Variance floor.** One may replace $s_2(u)$ by $\max\{s_2(u), v_{\min}\}$ before the square root. This guarantees the lower bound in Eq. 16 for all inputs.

- **Variance clamping.** One may replace $s_2(u)$ by $\mathrm{clip}(s_2(u), v_{\min}, v_{\max})$. This guarantees both inequalities in Eq. 16 for all inputs, so the constants in Eq. 17 hold deterministically.

Even without explicit clipping, Eq. 17 holds on any subset of states where the empirical variance stays within $[v_{\min}, v_{\max}]$. In particular, the upper bound $c_{\max} \leq 1$ always holds, while the lower bound is controlled by how far the monitored variance stays away from 0 relative to $\varepsilon$.

### A.2. Proof of Lemma 3.1

Using Eq. (1) and Eq. (2), $\tilde{h}_{t,b} = \Gamma_b \mathcal{S}(u_{t,b})$. Since $\|\Gamma_b\|_2 \leq \gamma_{\max}$ and $\|\Gamma_b^{-1}\|_2 \leq 1/\gamma_{\min}$, Eq. (3) yields $\gamma_{\min} c_{\min} \sqrt{d} \leq \|\tilde{h}_{t,b}\|_2 \leq \gamma_{\max} c_{\max} \sqrt{d}$, which gives Eq. (4). □

### A.3. Window indexing and deterministic assignment $j(b)$

This appendix fixes the window list $\{\mathcal{W}_j\}_{j=1}^{J}$ and the deterministic mapping $j(b)$ used in Sec. 3.2. The construction guarantees that the final window ends at block $B - 1$, while supporting overlap under stride $s < L$.

**Index set.** Blocks are indexed by integers $b \in \{0, 1, \ldots, B-1\}$.

**Window starts and ends with a forced final window.** Fix integers $L \geq 1$ (window length) and $s \geq 1$ (stride). Define the forced last start index

$$b_{\mathrm{last}} = \max\{0,\ B - L\}.$$

Let $J$ be the smallest integer such that $(J-1)s \geq b_{\mathrm{last}}$, equivalently

$$J = \left\lceil \frac{b_{\mathrm{last}}}{s} \right\rceil + 1.$$

For $j \in \{1, \ldots, J\}$, define the start index

$$b_j = \min\{(j-1)s,\ b_{\mathrm{last}}\},$$

and the end index

$$e_j = \min\{b_j + L - 1,\ B - 1\}.$$

The window is the contiguous index set

$$\mathcal{W}_j = \{b_j,\ b_j + 1,\ \ldots,\ e_j\}.$$

By construction, $b_J = b_{\mathrm{last}}$ and $e_J = B - 1$, so the final window is always aligned to end at the last block. When $B \geq L$, the final window has length exactly $L$. When $B < L$, we have $b_{\mathrm{last}} = 0$ and the single window $\mathcal{W}_1 = \{0, \ldots, B-1\}$ covers the full depth axis.

**Deterministic assignment** $j(b)$**.** When $s < L$, a block index can belong to multiple windows. We assign each block to the latest started window among those that contain it. For each $b \in \{0, \ldots, B-1\}$, define

$$j(b) = \max\{j \in \{1, \ldots, J\} : b \in \mathcal{W}_j\}.$$

Because $\mathcal{W}_J$ contains $B-1$ and windows are contiguous, the set is nonempty for every $b$ and the maximum exists and is unique.

A closed form uses the condition $b \le e_j$:

$$b \le e_j \iff b \le b_j + L - 1 \iff b_j \ge b - (L-1).$$

Since $b_j = \min\{(j-1)s, b_{\text{last}}\}$ is nondecreasing in $j$, the latest window containing $b$ is obtained by taking the largest admissible start, yielding

$$j(b) = \min\left\{J, \left\lceil \frac{\max\{0, b-(L-1)\}}{s} \right\rceil + 1\right\}.$$

This rule reduces to $j(b) = \min\{J, \lfloor b/s \rfloor + 1\}$ when $L$ is large enough that membership is implied by $b_j \le b$, and it correctly handles overlap by enforcing $b \in \mathcal{W}_{j(b)}$.

**Basic properties.** The assignment $j(b)$ is nondecreasing in $b$ and increments only when $b$ crosses a new eligible window start under the membership constraint. In particular, for each $b$, the assigned window satisfies

$$b_{j(b)} \le b \le e_{j(b)}.$$

**Step transition logic for transports.** The transport selector in Sec. 3.3.1 uses $j(b)$ and $j(b+1)$. A window switch occurs at depth step $b \to b+1$ exactly when $j(b+1) \ne j(b)$.

**Reference pseudocode.** The following pseudocode matches the equations above.

```
# Inputs: B, L, s
b_last = max(0, B - L)
J = ceil(b_last / s) + 1
for j in {1,...,J}:
    b_j = min((j-1)*s, b_last)
    e_j = min(b_j + L - 1, B - 1)
    W_j = {b_j, ..., e_j}

# Deterministic assignment (latest window that contains b):
for b in {0,...,B-1}:
    j_of_b = min(J, ceil(max(0, b-(L-1)) / s) + 1)
```

**Small worked example.** Take $B = 10$, $L = 4$, $s = 2$. Then $b_{\text{last}} = 6$ and $J = \lceil 6/2 \rceil + 1 = 4$. Starts are $b_j \in \{0, 2, 4, 6\}$ and endpoints are $e_j \in \{3, 5, 7, 9\}$, so $\mathcal{W}_4 = \{6, 7, 8, 9\}$ ends at $B-1$. The assignment yields $j(b) = 1$ for $b \in \{0, 1, 2, 3\}$, $j(b) = 2$ for $b \in \{4, 5\}$, $j(b) = 3$ for $b \in \{6, 7\}$, and $j(b) = 4$ for $b \in \{8, 9\}$.

### A.4. Competitor construction and token mask $M$

This appendix fixes the precise construction of the top token $\hat{y}_{t,b}$, the competitor set $\mathcal{C}_{t,b}$ used in Eq. 5, and the predefined token mask $M$ that restricts which token positions contribute directions to $\mathcal{A}_j$.

**Logits used for ranking.** At each monitored boundary and depth $b$, we compute logits

$$\ell_{t,b} = W h_{t,b} \in \mathbb{R}^V.$$

When bias-centering is used, $h_{t,b} = \tilde{h}_{t,b} + \beta_b$ (Eq. 1), so the same logits can be written as $\ell_{t,b} = W(\tilde{h}_{t,b} + \beta_b)$. All ranking operations below use these logits and do not apply temperature rescaling, logit clipping, or post-processing.

**Top token and tie breaking.** Define the top token index

$$\widehat{y}_{t,b} \in \arg\max_{y \in \{1,\ldots,V\}} \ell_{t,b}[y].$$

If the argmax is not unique, break ties deterministically by selecting the smallest index:

$$\widehat{y}_{t,b} = \min\left\{y : \ell_{t,b}[y] = \max_{y'} \ell_{t,b}[y']\right\}.$$

**Competitor set.** Let $\pi_{t,b}$ be a deterministic ordering of token indices obtained by sorting logits in non-increasing order, with ties broken by the same smallest-index rule. Thus,

$$\ell_{t,b}[\pi_{t,b}(1)] \geq \ell_{t,b}[\pi_{t,b}(2)] \geq \cdots \geq \ell_{t,b}[\pi_{t,b}(V)].$$

Then $\pi_{t,b}(1) = \widehat{y}_{t,b}$. Fix an integer $K \geq 1$. The competitor set is the next $K$ distinct indices after the top token:

$$\mathcal{C}_{t,b} = \{\pi_{t,b}(2), \pi_{t,b}(3), \ldots, \pi_{t,b}(K+1)\}. \tag{21}$$

If $K + 1 > V$, we take all remaining indices.

**Competitor difference directions.** With readout rows $\{w_y\}_{y=1}^V$ and $(t, b)$ fixed, define the competitor directions

$$a_{t,b,y} = w_{\widehat{y}_{t,b}} - w_y, \qquad y \in \mathcal{C}_{t,b}.$$

These are exactly the directions collected into $\mathcal{A}_j$ in Eq. 5.

**Token mask $M$.** For a given sequence of length $n$, define a binary mask

$$M \in \{0,1\}^n, \qquad M_t = 1 \iff t \text{ is eligible for competitor sampling.}$$

The masked restriction in the main text means that $(t, b)$ contributes directions only when $M_t = 1$. The mask is deterministic given the tokenization and sequence metadata, and is chosen to exclude positions whose logits are either not semantically meaningful or are dominated by formatting and control tokens.

**Default admissible positions.** Let $\mathcal{S}$ be the set of tokenizer special tokens together with any model-specific role or delimiter tokens, and let $x_{1:n}$ be the tokenized prompt-plus-generation sequence. A default mask used throughout is

$$M_t = \mathbf{1}\Big[x_t \notin \mathcal{S}\Big] \cdot \mathbf{1}\Big[t \leq n_{\text{eff}}\Big]. \tag{22}$$

where $n_{\text{eff}}$ is the effective sequence length after truncation and padding removal.

**Prompt and response restriction.** When a prompt prefix is present, we optionally restrict competitor sampling to a contiguous token span $t \in [t_{\min}, t_{\max}]$ to avoid control-heavy regions. In that case,

$$M_t \leftarrow M_t \cdot \mathbf{1}[t_{\min} \leq t \leq t_{\max}]. \tag{23}$$

A common choice is to set $t_{\min}$ to the first content token after the last role or delimiter token, and set $t_{\max}$ to the last generated content token before any stop token.

**Degenerate masked sets.** If $\{t : M_t = 1\}$ is empty for a sample, then $\mathcal{A}_j$ receives no directions from that sample at any depth. This case is handled by the subspace fitting fallback described in Appendix A.5.

### A.5. Subspace fitting details and degenerate windows

This appendix specifies how the sampled competitor directions in $\mathcal{A}_j$ (Eq. 5) are converted into the window matrix $D_j$ and the orthonormal basis $U_j$ (Eq. 6). It also defines a deterministic fallback when $\mathcal{A}_j$ is empty or numerically degenerate.

**Direction pool for window $j$.** Let
$$\mathcal{A}_j = \{a_{t,b,y} = w_{\widehat{y}_{t,b}} - w_y : b \in \mathcal{W}_j, \, y \in \mathcal{C}_{t,b}, \, M_t = 1\} \subset \mathbb{R}^d$$
be the masked competitor direction set for window $j$.

**Row normalization and clipping.** For each $a \in \mathcal{A}_j$, form a normalized direction
$$\bar{a} = \frac{a}{\|a\|_2 + \varepsilon_a}, \tag{24}$$
with a fixed $\varepsilon_a > 0$. This removes scale variability induced by readout-row norms and makes the SVD capture dominant angular structure rather than magnitude outliers. Optionally, we discard $a$ when $\|a\|_2 < \tau_a$ for a fixed threshold $\tau_a > 0$, since such directions can be dominated by numerical noise. We choose $\tau_a$ on the same numerical scale as $\varepsilon_a$.

**Capping and sampling.** Let $N_j = |\mathcal{A}_j|$ be the number of eligible directions for window $j$. Fix a cap $\mathrm{cap} \geq 1$. If $N_j \leq \mathrm{cap}$, use all normalized directions. If $N_j > \mathrm{cap}$, sample a subset of size $\mathrm{cap}$ without replacement using a deterministic seed
$$\mathrm{seed} = \mathrm{hash}(\mathrm{sample\ id}, j, \mathrm{seed}_0),$$
with a fixed $\mathrm{seed}_0$, ensuring reproducibility across runs. Denote the resulting ordered list by $(\bar{a}^{(1)}, \ldots, \bar{a}^{(M_j)})$, with $M_j \leq \mathrm{cap}$.

**Window matrix.** Stack the sampled normalized directions as rows:
$$D_j = \begin{bmatrix} (\bar{a}^{(1)})^\top \\ \vdots \\ (\bar{a}^{(M_j)})^\top \end{bmatrix} \in \mathbb{R}^{M_j \times d}. \tag{25}$$
This is the matrix whose right singular vectors define the readout-aligned window subspace.

**SVD and basis extraction.** Compute a thin SVD
$$D_j = U_j^{(D)} \Sigma_j V_j^\top, \qquad \Sigma_j = \mathrm{diag}(\sigma_{j,1}, \ldots, \sigma_{j,r}), \quad \sigma_{j,1} \geq \cdots \geq \sigma_{j,r} > 0,$$
where $r = \mathrm{rank}(D_j) \leq \min\{M_j, d\}$. Write the right singular vectors as $V_j = [v_{j,1}, \ldots, v_{j,d}]$. Extract the top-$k$ right singular vectors
$$\widetilde{U}_j = [v_{j,1}, \ldots, v_{j,k}] \in \mathbb{R}^{d \times k}. \tag{26}$$
To improve numerical stability, we re-orthonormalize
$$\widetilde{U}_j = Q_j R_j \quad (\text{thin QR}), \qquad U_j = Q_j, \tag{27}$$
so that $U_j^\top U_j = I_k$ up to numerical precision. If $r < k$, we extract only the available $r$ singular vectors and complete the basis with a deterministic orthonormal complement as described below.

**Degenerate windows and deterministic fallback.** A window is degenerate if $M_j = 0$ after masking and filtering. Otherwise, it is degenerate if $\mathrm{rank}(D_j) = 0$. In this case, $D_j$ contains no usable direction signal, so $U_j$ cannot be identified from competitor geometry. We define a deterministic fallback basis $U_j^{\mathrm{fb}}$ as follows.

Fix once a global reference orthonormal matrix $G \in \mathbb{R}^{d \times d}$. One convenient choice is the identity $G = I_d$. Define
$$U_j^{\mathrm{fb}} = [g_1, \ldots, g_k], \tag{28}$$
where $G = [g_1, \ldots, g_d]$. Set $U_j = U_j^{\mathrm{fb}}$ for degenerate windows. This choice is deterministic and yields well-defined signatures when competitor geometry is unavailable.

**Rank-deficient completion.** When $0 < r < k$, let $U_{j,1:r}$ be the extracted $r$ right singular vectors. Complete to $k$ columns by projecting $G$ to the orthogonal complement of $\mathrm{span}(U_{j,1:r})$ and taking the first $(k-r)$ directions, followed by a thin QR:

$$\widehat{G} = (I - U_{j,1:r}U_{j,1:r}^\top)G, \qquad \widehat{U}_j = \begin{bmatrix} U_{j,1:r} & \widehat{G}_{:,1:(k-r)} \end{bmatrix}, \qquad U_j = \mathrm{qf}(\widehat{U}_j),$$

where $\mathrm{qf}(\cdot)$ denotes the $Q$ factor of a thin QR decomposition.

**Deterministic reproducibility.** All stochasticity can be removed by using the deterministic fallback $G$ and a deterministic subsampling order. When randomized subsampling is used for cap, the seed must be fixed and recorded, and the same seed must be used across all runs for comparability.

### A.6. Orthogonal transport, Procrustes optimality, and corner cases

This appendix formalizes the orthogonal transport used to align adjacent window frames (Sec. 3.3.1) and states its Procrustes optimality. It also records deterministic choices for corner cases where the overlap between adjacent subspaces is weak.

**Setup.** Let $U, V \in \mathbb{R}^{d \times k}$ have orthonormal columns, representing two window bases (e.g., $U = U_j$ and $V = U_{j+1}$). Consider the orthogonal group $O(k) = \{R \in \mathbb{R}^{k \times k} : R^\top R = I_k\}$.

**Transport definition.** Compute the compact SVD

$$V^\top U = P\Sigma Q^\top, \qquad P, Q \in \mathbb{R}^{k \times k} \text{ orthogonal}, \quad \Sigma = \mathrm{diag}(\sigma_1, \ldots, \sigma_k),\ 0 \leq \sigma_i \leq 1. \tag{29}$$

Define the transport

$$R^\star = PQ^\top \in O(k). \tag{30}$$

This is the orthogonal factor used in the main text as $R_{j \to j+1}$.

**Lemma A.3** (Procrustes optimality of the transport). *Let $U, V \in \mathbb{R}^{d \times k}$ have orthonormal columns and let $R^\star$ be defined by Eqs. 29–30. Then*

$$R^\star \in \arg\min_{R \in O(k)} \|U - VR\|_F, \tag{31}$$

*and equivalently*

$$R^\star \in \arg\max_{R \in O(k)} \mathrm{tr}(R^\top V^\top U). \tag{32}$$

*Moreover, the minimum value satisfies*

$$\min_{R \in O(k)} \|U - VR\|_F^2 = 2k - 2\sum_{i=1}^k \sigma_i, \tag{33}$$

*where $\{\sigma_i\}$ are the singular values of $V^\top U$.*

**Proof.** Expand

$$\|U - VR\|_F^2 = \mathrm{tr}(U^\top U) + \mathrm{tr}(R^\top V^\top V R) - 2\mathrm{tr}(R^\top V^\top U) = 2k - 2\mathrm{tr}(R^\top V^\top U),$$

so minimizing the Frobenius distance is equivalent to maximizing $\mathrm{tr}(R^\top V^\top U)$ over $R \in O(k)$, giving Eq. 32. Using the SVD $V^\top U = P\Sigma Q^\top$,

$$\mathrm{tr}(R^\top V^\top U) = \mathrm{tr}(R^\top P\Sigma Q^\top) = \mathrm{tr}((P^\top RQ)^\top \Sigma).$$

Let $S = P^\top RQ \in O(k)$. Then $\mathrm{tr}(S^\top \Sigma) = \sum_{i=1}^k \sigma_i S_{ii} \leq \sum_{i=1}^k \sigma_i$, since $|S_{ii}| \leq 1$ and $\sigma_i \geq 0$. Equality is achieved by $S = I_k$, which corresponds to $R = PQ^\top = R^\star$. Substituting the maximizing trace into the expansion yields Eq. 33.

**Interpretation.** The transport $R^\star$ is the closest orthogonal change of coordinates between the two $k$-frames in the least-squares sense. Applying $R^\star$ aligns coordinates from the $U$-frame to the $V$-frame while minimizing window-switch-induced rotation artifacts in $k$-space.

**Uniqueness and sign ambiguities.** If $V^\top U$ has distinct singular values and full rank, then $R^\star$ is unique. When singular values repeat, the SVD factors $P, Q$ are not unique, but any valid choice yields a transport that attains the same Procrustes optimum in Eq. 31. This non-uniqueness corresponds exactly to the within-subspace gauge freedom and does not affect gauge-invariant signatures.

**Weak overlap and numerically unstable cases.** When adjacent subspaces have weak overlap, some singular values of $V^\top U$ can be close to 0, and the Procrustes transport $R^\star = PQ^\top$ can become numerically unstable. We therefore use a deterministic reset rule based on the smallest overlap.

Fix a threshold $\tau_\sigma \in (0,1)$ and compute the compact SVD in Eq. 29. Let $\sigma_{\min} = \min_{i \in \{1,\ldots,k\}} \sigma_i$. We define the stabilized transport

$$R^{\text{stab}} = \begin{cases} R^\star, & \sigma_{\min} \geq \tau_\sigma, \\ I_k, & \sigma_{\min} < \tau_\sigma. \end{cases} \quad (34)$$

This rule treats a weak-overlap window switch as a complete frame reset. The resulting discontinuity is captured by the drift metrics in Sec. 3.3.3.

In the main text, $R_{j \to j+1}$ refers to $R^\star$ by default, with Eq. 34 applied deterministically when overlap is weak.

### A.7. Rotation-equivariant robust centering in $k$-space

This appendix fixes the per-step center $\mu_b \in \mathbb{R}^k$ used in Eq. 10 and records sufficient conditions under which the centered increment norm $\|\Delta p_{t,b}^c\|_2$ is invariant to within-window basis rotations, as required by Lemma 3.2.

**Problem statement.** For a fixed sample and depth step $b$, let $\mathcal{T}_b$ be the set of eligible token indices (those with $M_t = 1$). Given transported increments $\{\Delta p_{t,b} \in \mathbb{R}^k : t \in \mathcal{T}_b\}$, we seek a center map

$$\mu : (\mathbb{R}^k)^{|\mathcal{T}_b|} \to \mathbb{R}^k, \qquad \mu_b = \mu(\{\Delta p_{t,b}\}_{t \in \mathcal{T}_b}),$$

that is robust to outliers and equivariant under rotations in $\mathbb{R}^k$.

**Rotation equivariance.** A center rule $\mu(\cdot)$ is *rotation-equivariant* if, for every orthogonal $Q \in O(k)$,

$$\mu(\{Q^\top v_t\}_t) = Q^\top \mu(\{v_t\}_t). \quad (35)$$

If Eq. 35 holds, then the centered residuals rotate consistently:

$$(Q^\top v_t) - \mu(\{Q^\top v_t\}_t) = Q^\top(v_t - \mu(\{v_t\}_t)),$$

so Euclidean norms $\|v_t - \mu(\cdot)\|_2$ are preserved.

**Default choice: masked coordinate-wise median.** We choose $\mu_b$ as the coordinate-wise median over the eligible set:

$$(\mu_b)_i = \text{median}\Big(\{(\Delta p_{t,b})_i : t \in \mathcal{T}_b\}\Big), \qquad i = 1, \ldots, k. \quad (36)$$

We use the convention that masked entries are ignored and an all-masked set returns $\mu_b = 0$.

**Computation via Weiszfeld iterations.** When $|\mathcal{T}_b| \geq 1$, we compute an approximate geometric median by Weiszfeld's algorithm. Initialize $\mu_b^{(0)}$ as the Euclidean mean of $\{\Delta p_{t,b}\}_{t \in \mathcal{T}_b}$.

For $r = 0, 1, \ldots, R - 1$, update

$$\mu_b^{(r+1)} = \frac{\sum_{t \in \mathcal{T}_b} w_t^{(r)} \Delta p_{t,b}}{\sum_{t \in \mathcal{T}_b} w_t^{(r)}}, \qquad w_t^{(r)} = \frac{1}{\|\Delta p_{t,b} - \mu_b^{(r)}\|_2 + \varepsilon_\mu}, \quad (37)$$

with fixed $\varepsilon_\mu > 0$ for numerical stability. Set $\mu_b = \mu_b^{(R)}$. If any iterate coincides with an input point (within tolerance), we return that point, which is a valid minimizer.

**Empty mask.** If $\mathcal{T}_b = \emptyset$, define $\mu_b = 0 \in \mathbb{R}^k$ and skip centering for that step.

**Alternative equivariant centers.** The following center rules satisfy the rotation-equivariance property in Eq. 35 because they depend only on Euclidean distances and are invariant under orthogonal changes of coordinates:

- **Euclidean mean:** $\mu_b = \frac{1}{|\mathcal{T}_b|} \sum_{t \in \mathcal{T}_b} \Delta p_{t,b}$.
- **Geometric median:** any $\mu_b \in \arg\min_u \sum_{t \in \mathcal{T}_b} \|\Delta p_{t,b} - u\|_2$.
- **Huber-type radial center:** any $\mu_b \in \arg\min_u \sum_{t \in \mathcal{T}_b} \psi(\|\Delta p_{t,b} - u\|_2)$ for a radial loss $\psi$.
- **Distance-based trimmed mean:** discard a fixed fraction of points farthest from the mean (or another equivariant pilot) using Euclidean distances, then average the remainder.

In contrast, the coordinate-wise median (used in our implementation) is robust but is not rotation-equivariant for general $Q \in O(k)$.

**Invariance of centered increment norms.** Suppose a within-window basis change induces an orthogonal coordinate transform in $k$-space, so that $\Delta p'_{t,b} = Q^\top \Delta p_{t,b}$ for some $Q \in O(k)$ and all eligible $t$ at step $b$. If the chosen center rule is rotation-equivariant (Eq. 35), then centering commutes with the rotation:

$$\Delta p_{t,b}^{c'} = \Delta p'_{t,b} - \mu(\{\Delta p'_{t,b}\}_t) = Q^\top \Delta p_{t,b} - Q^\top \mu(\{\Delta p_{t,b}\}_t) = Q^\top \Delta p_{t,b}^c,$$

and therefore the centered norms are invariant,

$$\|\Delta p_{t,b}^{c'}\|_2 = \|\Delta p_{t,b}^c\|_2.$$

This is the sufficient condition used in Lemma 3.2. In our experiments we center by the masked coordinate-wise median for robustness; this choice does not satisfy Eq. 35 in general, but it matches the implementation and the reported results.

## A.8. Drift metric properties and simple bounds

This appendix records basic properties of the drift quantities in Sec. 3.3.3. We relate the projector spectral drift

$$d_G(U, V) = \|UU^\top - VV^\top\|_2$$

to principal angles, and we give simple bounds for the anchor-coupled drift $\chi_{t,j}$ in Eq. 14.

**Projectors and principal angles.** Let $U, V \in \mathbb{R}^{d \times k}$ have orthonormal columns, and let $P_U = UU^\top$ and $P_V = VV^\top$ be the associated orthogonal projectors. Let $\theta_1 \leq \cdots \leq \theta_k \in [0, \pi/2]$ be the principal angles between $\text{span}(U)$ and $\text{span}(V)$. Equivalently, if $U^\top V$ has singular values $\sigma_1 \geq \cdots \geq \sigma_k$ in $[0, 1]$, then

$$\sigma_i = \cos(\theta_i).$$

**Lemma A.4** (Spectral projector drift equals $\sin(\theta_{\max})$). *With notation above,*

$$\|P_U - P_V\|_2 = \sin(\theta_{\max}), \qquad \theta_{\max} = \theta_k. \tag{38}$$

*In particular, $0 \leq d_G(U, V) \leq 1$, and $d_G(U, V) = 0$ if and only if the two subspaces coincide.*

**Proof sketch.** This is a standard identity for orthogonal projectors onto $k$-dimensional subspaces. One route is to use a principal-angle coordinate system in which $P_U - P_V$ decomposes into $k$ independent $2 \times 2$ blocks with eigenvalues $\pm \sin(\theta_i)$, so the spectral norm equals $\max_i \sin(\theta_i)$.

**Useful corollaries.** The following elementary bounds are often convenient.

- **Frobenius drift:** $\|P_U - P_V\|_F^2 = 2 \sum_{i=1}^k \sin^2(\theta_i)$.
- **Trace overlap:** $\text{tr}(P_U P_V) = \|U^\top V\|_F^2 = \sum_{i=1}^k \cos^2(\theta_i)$.
- **Overlap lower bound:** if $\sigma_{\min}(U^\top V) \geq \alpha$, then $d_G(U, V) \leq \sqrt{1 - \alpha^2}$.

**Anchor-coupled drift and bounds.** Recall Eq. 14:

$$\chi_{t,j} = \frac{\|(P_{j+1} - P_j)\tilde{h}_{t,b_j^\star}\|_2}{\|\tilde{h}_{t,b_j^\star}\|_2 + \varepsilon}, \qquad P_j = U_j U_j^\top.$$

We first record a deterministic rule for the anchor $b_j^\star$, then state simple bounds.

**Anchor selection.** We use the window start as the anchor,

$$b_j^\star = b_j, \tag{39}$$

with $b_j = (j-1)s$ from Appendix A.3. This rule is deterministic and ensures each anchor belongs to its own window.

**Lemma A.5** (Simple bounds for $\chi_{t,j}$). *For any nonzero $v \in \mathbb{R}^d$ and any two orthogonal projectors $P, Q$,*

$$\frac{\|(P-Q)v\|_2}{\|v\|_2 + \varepsilon} \leq \|P - Q\|_2.$$

*Consequently, for $\chi_{t,j}$,*

$$0 \leq \chi_{t,j} \leq d_G(U_j, U_{j+1}), \tag{40}$$

*and also*

$$\chi_{t,j} \leq 1. \tag{41}$$

**Proof.** By submultiplicativity,

$$\|(P-Q)v\|_2 \leq \|P-Q\|_2 \|v\|_2,$$

and since $\|v\|_2/(\|v\|_2 + \varepsilon) \leq 1$, the displayed inequality follows. Apply it with $P = P_{j+1}$ and $Q = P_j$ to obtain Eq. 40. Eq. 41 follows from $\|P_{j+1} - P_j\|_2 \leq 1$ for orthogonal projectors onto equal-dimensional subspaces.

**Interpretation.** The geometry-only drift $d_G(U_j, U_{j+1})$ measures the worst-case mismatch direction between the two adjacent $k$-subspaces via the maximum principal angle. The anchor-coupled drift $\chi_{t,j}$ measures how much of that subspace change is expressed along the actual visited anchor state $\tilde{h}_{t,b_j^\star}$, and the bound $\chi_{t,j} \leq d_G(U_j, U_{j+1})$ ensures the state-coupled drift cannot exceed the geometry-only drift.

### A.9. Refinement via Readout-Aligned Step Clamping

This appendix defines the single-block refinement operator used after the validator localizes a culprit depth event. Implementation choices such as competitor selection, basis caching, calibration span, and stopping are documented separately in Appendix B.11.

**Target block and intervention position.** Given a flagged sample, the validator identifies a culprit event index $j^\star$ and its associated coordinates $(t_0, b_0) = (t_{j^\star}, b_{j^\star})$ as described in Sec. 3.5. Refinement keeps the prefix $x_{1:t_0}$ fixed and regenerates the remaining suffix while modifying the internal update only at block $b_0$.

**Local readout-aligned subspace at block $b_0$.** Let $W \in \mathbb{R}^{V \times d}$ be the readout matrix with rows $\{w_y\}_{y=1}^V$. At the monitored boundary of block $b_0$, let $h^{\text{in}}, h^{\text{out}} \in \mathbb{R}^d$ denote the block input and output at the current decoding position (using the same monitored state convention as in Sec. 3.1). A local orthonormal basis $U \in \mathbb{R}^{d \times k}$ is fitted from readout-row difference directions between the current top token and a set of competitors, with $k$ denoting the subspace dimension and $K$ the number of competitors (Appendix B.11). We assume

$$U^\top U = I_k. \tag{42}$$

**$k$-space coordinates and transported step.** Project the states to $k$-space

$$p^{\text{in}} = U^\top h^{\text{in}} \in \mathbb{R}^k, \qquad p^{\text{out}} = U^\top h^{\text{out}} \in \mathbb{R}^k. \tag{43}$$

To allow distinct bases on the two sides of the block, introduce an orthogonal transport $R \in \mathbb{R}^{k \times k}$. The transported step and its norm are

$$\Delta p = p^{\text{out}} - R p^{\text{in}} \in \mathbb{R}^k, \qquad s = \|\Delta p\|_2. \tag{44}$$

In our default setting we use the same basis on both sides, so $R = I_k$.

**Reference scale from the fixed prefix.** A calibration pass on the fixed prefix yields a robust reference step scale $s_{\text{ref}} > 0$ computed from a short span of prefix positions. The aggregation rule and span selection are specified in Appendix B.11.

**Upper clamp and subspace-only rewrite.** Given a clamp ratio $\alpha > 1$, define

$$\lambda = \min\left(1, \frac{\alpha\, s_{\text{ref}}}{s + \varepsilon}\right), \qquad \Delta p' = \lambda\, \Delta p, \qquad p^{\text{out}\prime} = R\, p^{\text{in}} + \Delta p'. \tag{45}$$

We rewrite only the component in $\text{span}(U)$ and preserve the orthogonal complement:

$$h^{\text{out}\prime} = h^{\text{out}} + U\bigl(p^{\text{out}\prime} - p^{\text{out}}\bigr). \tag{46}$$

The intervention replaces $h^{\text{out}}$ by $h^{\text{out}\prime}$ only at block $b_0$ and only at the current decoding position.

**Scope.** The operator applies only for suffix generation after position $t_0$ and only at a single block $b_0$. All other blocks and all earlier prefix positions remain unchanged.

# B. Implementation Details

## B.1. Deterministic conventions and numerical constants

All runs follow fixed conventions for reproducibility. We fix a single random seed per run and apply it consistently to Python, NumPy, and PyTorch. We also use a single safeguard constant $\varepsilon_{\text{num}} > 0$ in every denominator involving an $\ell_2$ norm in the flow pipeline and validator features.

**Boundary normalization constant.** The monitored boundary applies the model's pretrained normalization module, either LayerNorm or RMSNorm depending on the architecture. We denote its built-in epsilon by $\varepsilon_{\text{bn}}$ and take $\varepsilon_{\text{bn}}$ directly from the pretrained configuration. All statements that use a normalization band in Lemma 3.1 and Appendix A.1 refer to this boundary normalization and its $\varepsilon_{\text{bn}}$.

## B.2. Token masks and validity priority

Each sequence uses two binary masks: the tokenizer attention mask $a_{n,t}$ for token validity, and an eligibility mask $m_{n,t}$ that selects positions used in aggregation and centering. The attention mask has strict priority, so any position with $a_{n,t} = 0$ is treated as invalid. All pooling and centering operate only on tokens with $a_{n,t} = 1$ and $m_{n,t} = 1$.

## B.3. Robust masked aggregation over depth

We aggregate per-step ratio features over depth into token-level summaries using a masked median. For each sample $n$ and token $t$, we take the median of $\{r_{n,t,b}\}$ over depth steps that are valid under the token masks and lie within the effective monitored range ($b \leq B_{n,\text{eff}}$). If no valid depth step exists for a token, we set its summary to $0$. This robust reduction stabilizes summaries under outlier steps and variable effective depths.

## B.4. Transported increments and robust masked centering

For each sample $n$ and depth step $b$, we center transported increments $\Delta p_{n,t,b} \in \mathbb{R}^k$ across eligible tokens to remove token-shared shifts, so pooled motion reflects token-specific deviations. Using the eligible set $\mathcal{T}_n = \{t : a_{n,t} = 1,\ m_{n,t} = 1\}$, we subtract a robust masked center in $k$-space computed by a coordinate-wise median:

$$\mu_{n,b}[i] = \text{median}\bigl(\{\Delta p_{n,t,b}[i] : t \in \mathcal{T}_n\}\bigr), \quad i \in \{1, \ldots, k\}, \qquad \Delta \bar{p}_{n,t,b} = \Delta p_{n,t,b} - \mu_{n,b},$$

with the convention $\mu_{n,b} = 0$ when $\mathcal{T}_n = \emptyset$. This is the centering used in all experiments. Note that coordinate-wise medians are robust but are not rotation-equivariant for general within-window orthogonal basis changes; using a rotation-equivariant center (e.g., Euclidean mean or geometric median) is a drop-in alternative if strict equivariance is desired.

## B.5. Safe normalization for ratio features

All ratio features use a shared numerical safeguard $\varepsilon_{\text{num}} > 0$ to avoid instability when norms are small. Whenever a norm appears in a denominator, we use $\|v\|_2 + \varepsilon_{\text{num}}$. This rule is applied to direction normalization in $k$-space and to all norm-based ratios involving projected updates or residual terms.

### B.6. Moving subspace and competitor hyperparameters

Unless noted otherwise, we use window length $L = 8$ and stride $s = 4$ for the moving subspace assignment in Sec. 3.2. For competitor directions and subspace fitting, we use $K = 32$ competitors and fit a $k = 16$ basis in the primary detection experiments, matching the refinement setting in Appendix B.11.

### B.7. Quadrature and Jacobian vector products

Path-integrated updates use Jacobian vector products of the boundary normalization map at depth $b+1$. We compute these products via automatic differentiation without forming explicit Jacobians. For the integral over $\alpha \in [0, 1]$, we use a fixed three-node Simpson-type rule with nodes $\{0, 0.5, 1\}$.

### B.8. Validator inputs

Each sample provides a depthwise feature grid with token validity. We represent one file payload as

$$x_{\text{grid}} \in \mathbb{R}^{N \times B_{\text{eff}} \times T \times F},$$

with $N$ samples, $B_{\text{eff}}$ monitored depth steps, token length $T$, and feature dimension $F$. Token validity follows the tokenizer attention mask and is applied uniformly across depth steps by masking invalid token positions before any reduction.

To form a sequence input for the validator, we pad within a minibatch to a common shape $(B_{\max}, T_{\max})$ and linearize the depth–token grid into a single event axis using a fixed index order. This yields an event tensor

$$\text{evt\_x} \in \mathbb{R}^{M \times L \times F}, \qquad L = B_{\max} T_{\max},$$

with a matching binary mask $\text{evt\_valid} \in \{0, 1\}^{M \times L}$ indicating valid events after padding. All downstream computations treat masked events as absent.

### B.9. Validator architecture and masked pooling

The validator takes the event sequence $\text{evt\_x}$ together with $\text{evt\_valid}$. We first apply a feature-wise LayerNorm to stabilize scale across feature channels, then map each event vector to an embedding through a lightweight MLP. A GRU processes the resulting event sequence, and the final decision aggregates per-event logits through a mask-aware pooling operator that ignores invalid events.

We use either max pooling or logsumexp pooling over valid event positions. Pooling is applied after the GRU so that event ordering can influence the hidden dynamics prior to aggregation.

**Default hyperparameters.** Unless noted otherwise, the event encoder uses a two-layer MLP with hidden size 256, embedding size $d_e = 128$, and dropout 0.1, followed by an output LayerNorm. The GRU uses hidden size $d_h = 256$ with a single recurrent layer.

### B.10. Training and evaluation procedure

We train and evaluate validators independently for each (task, base LLM) pair. For each pair, we split the corresponding serialized flow files into train and test with ratio 8:2, and all reported numbers use the held out test split. During both training and testing, we keep only samples whose labels fall in $\{0, 1\}$. Batches that contain no remaining binary labeled samples after filtering are skipped and counted. We optimize all validator parameters with AdamW using learning rate $3 \times 10^{-5}$ and weight decay $10^{-2}$ for 300 epochs. Optional gradient clipping uses global norm threshold 1.0. Optional mixed precision uses autocast with a GradScaler, controlled by a run flag. To address class imbalance, we use weighted binary cross entropy on logits. We scan the training split once, count binary labels, and set the positive class weight to the ratio of negative to positive counts, falling back to weight 1.0 when no positives appear. For evaluation, we convert logits to probabilities with a sigmoid. Accuracy uses fixed threshold 0.5. AUROC uses predicted probabilities and binary labels.

### B.11. Flow guided refinement intervention

We apply refinement only to samples labeled as hallucinations ($y = 1$) under the evaluation protocol used in Table 2. For each selected sample, the extractor provides an event sequence with per-event scores and coordinates $(t_j, b_j)$ over token index and depth. We select the culprit event by masked maximization over valid events and use its coordinates $(t_0, b_0)$ to determine the intervention token position and the single targeted block.

Refinement keeps the prefix $x_{1:t_0}$ fixed and regenerates the remaining suffix while intervening at exactly one transformer block $b_0$. The intervention rewrites only the component inside a readout-aligned $k$-dimensional subspace and preserves the orthogonal complement. We use a fixed competitor count $K$ and subspace dimension $k$, and we freeze the competitor set and fitted basis after calibration.

We estimate a reference step scale $s_{\text{ref}}$ from the fixed prefix by aggregating transported step norms over a bounded span. We set $s_{\text{ref}}$ to the masked median of $\|\Delta p\|_2$ over the first $N_{\text{cal}}$ valid prefix steps (default $N_{\text{cal}} = 64$). During suffix generation, if the transported step norm exceeds an upper band relative to $s_{\text{ref}}$, we apply a shrink-only update that preserves direction in the transported coordinates.

Unless noted otherwise, we use $K = 32$ competitors to fit a $k = 16$ readout-aligned basis, estimate $s_{\text{ref}}$ from up to 64 prefix steps, and use an upper-band clamp ratio $\alpha = 1.05$.

### B.12. External judge protocol for hallucination labels

We obtain binary hallucination labels from a ChatGPT based external judge conditioned on the same context and the base model output. We use a fixed prompt template across all tasks and base LLMs, and we use deterministic decoding for the judge. The judge emits labels in $\{-1, 0, 1\}$, and we keep only labels in $\{0, 1\}$ for training and evaluation.

## C. Flow Analyze Details

This appendix reports the supporting statistics for Each task on HaluEval, focusing on three views per model: (i) top group dominance (top1 fraction), (ii) group magnitude via Grad times Input group mass, and (iii) hotspot depth distribution. We also summarize label aligned and error mode aligned behaviors.

### C.1. QA Task

This section summarizes how the flow signatures behave on HaluEval QA and why the validator succeeds or fails across model families. The most consistent cue is a depth localized gate in the transported coordinates $p_{t,b}$. Predicted positives tend to collapse into a narrow depth band where multiple group masses rise together, while missed hallucinations more often remain in a late, diffuse regime that looks similar to non hallucination. Module top1 dominance can shift by label for some models, but it is less reliable than the depth gate and the coupled mass pattern.

**Overall pattern.** Across models, QA predicted positives frequently form a single depth gate with strong multi group co activation.

- **Depth gate drives prediction.** Predicted positives concentrate at one shallow or late depth index, depending on the base model.
- **TP and FP overlap in magnitude.** When the gate activates, group mass often saturates across motion, attention, MLP, competitor, and drift, so $TP$ and $FP$ look similar in magnitude.
- **FN looks TN like.** Missed hallucinations typically stay in a late, diffuse regime where hotspot depth and dominance resemble $TN$.

#### C.1.1. QWEN2.5

**Key pattern.** Predicted positives collapse to an early gate at depth $b = 1$. Within $y = 1$, detected hallucinations deviate from pure motion dominance, with higher attention and drift shares. Missed hallucinations remain motion dominated and resemble $TN$.

**Operating point.** Hallucination prevalence: 15.5%. Predicted positive rate: 23.4%. Precision: 38.8%. Recall: 58.4%.

**Depth localization.** The early gate dominates prediction: $TP$ at $b = 1$ is 100.0% and $FP$ at $b = 1$ is 99.6%. $TN$ hotspots spread across mid to late depth with representative peaks at $b = 18$ (12.7% of $TN$), $b = 19$ (11.6%), and $b = 20$ (9.0%). $FN$ has only a small early gate share at $b = 1$ (8.7% of $FN$) and otherwise stays late and diffuse.

**Dominance and mass.** Motion top1 drops from $95.4\%$ in $y = 0$ to $87.5\%$ in $y = 1$, while attention rises from $3.1\%$ to $7.2\%$ and drift rises from $0.3\%$ to $2.0\%$. Within $y = 1$, $TP$ further reduces motion dominance (motion $79.2\%$, attention $12.4\%$, drift $3.4\%$), whereas $FN$ stays motion dominant ($99.2\%$), close to $TN$ (motion $98.8\%$). Group mass separates mainly by prediction: predicted positives sit near $\approx 0.20$ across groups, while predicted negatives remain small ($TN \approx 0.013$, $FN \approx 0.024$).

### C.1.2. GEMMA2

**Key pattern.** A strong early gate at depth $b = 0$ snaps predicted positives into a single spike. Multi group mass saturates at the gate, producing heavy overlap between $TP$ and $FP$. Module dominance carries little signal.

**Operating point.** Hallucination prevalence: $15.0\%$. Predicted positive rate: $41.1\%$. Precision: $22.2\%$. Recall: $60.8\%$.

**Depth localization.** The gate dominates prediction: $TP$ at $b = 0$ is $100.0\%$ and $FP$ at $b = 0$ is $99.6\%$. By label, $b = 0$ accounts for $67.9\%$ of $y = 1$ and $44.0\%$ of $y = 0$, so early depth correlates with label but does not isolate it. $FN$ has a weaker early gate share ($b = 0$ at $18.1\%$ of $FN$) and keeps a late tail.

**Dominance and mass.** Top1 dominance stays near motion saturation (motion $99.6\%$ in $y = 1$, $99.5\%$ in $y = 0$). Group mass follows prediction: predicted positives saturate at large values (for example, motion $\approx 0.43$) while $TN$ remains smaller (motion $\approx 0.046$), and $FP$ closely matches $TP$ across groups.

### C.1.3. PHI 3

**Key pattern.** Phi 3 behaves like a depth and token gate. Predicted positives concentrate at depth $b = 0$ and the earliest tokens, while missed hallucinations drift toward later depth and later token hotspots closer to $TN$.

**Operating point.** Hallucination prevalence: $13.0\%$. Predicted positive rate: $63.2\%$. Precision: $14.8\%$. Recall: $71.9\%$.

**Depth and token localization.** Depth gating is absolute for prediction: $TP$ at $b = 0$ is $100.0\%$ and $FP$ at $b = 0$ is $100.0\%$. By label, $b = 0$ covers $84.6\%$ of $y = 1$, while $TN$ still contains a substantial $b = 0$ portion ($30.7\%$ of $TN$) and then spreads broadly into later depths. $FN$ splits, with $b = 0$ at $45.1\%$ of $FN$ and a strong late tail into the high 20s. Token hotspots mirror the gate: predicted positives concentrate at token 0 and 1 ($TP$: token 0 at $55.5\%$, token 1 at $39.6\%$; $FP$: token 0 at $52.7\%$, token 1 at $43.4\%$), while $TN$ peaks later (token 2 at $25.7\%$) and $FN$ also peaks at token 2 ($40.8\%$).

**Dominance and mass.** Module dominance does not separate quadrants, motion top1 is $100.0\%$ in all quadrants. Group mass appears quantized by prediction (predicted positives near $\approx 0.20$, $FN$ near $\approx 0.09$, $TN$ near $\approx 0.061$), so the depth and token gate description is more stable than label level dominance.

### C.1.4. LLAMA3

**Key pattern.** The decisive gate shifts late. Predicted positives concentrate around a sharp spike near depth $b = 21$, and the same spike often appears in $FP$, limiting label recoverability from depth alone.

**Operating point.** Hallucination prevalence: $15.1\%$. Predicted positive rate: $36.6\%$. Precision: $24.2\%$. Recall: $58.6\%$.

**Depth localization.** Predicted positives concentrate at $b = 21$ ($TP$: $62.1\%$ at $b = 21$, $FP$: $57.0\%$ at $b = 21$). By label, $b = 21$ accounts for $42.1\%$ of $y = 1$ and $23.7\%$ of $y = 0$, which produces a label shift with strong $FP$ overlap.

**Dominance and mass.** Top1 dominance is saturated for $y = 1$ (motion $100.0\%$), and $y = 0$ contains only tiny non motion shares (attention $0.41\%$, competitor $0.06\%$). Group mass follows prediction, with predicted positives near $\approx 0.20$ across groups and predicted negatives smaller ($TN \approx 0.025$, $FN \approx 0.043$).

### C.1.5. MISTRAL

**Key pattern.** Mistral forms a late depth spike gate in the $b = 20$ to $22$ neighborhood. When the spike appears, predicted positives rise, but $FP$ shares the same spike. Missed hallucinations remain diffuse and $TN$ like.

**Operating point.** Hallucination prevalence: $14.1\%$. Predicted positive rate: $51.6\%$. Precision: $16.9\%$. Recall: $61.9\%$.

**Depth localization.** Predicted positives concentrate in late depth: $TP$ peaks at $b = 21$ ($28.7\%$ of $TP$) and $b = 20$ ($17.2\%$), and $FP$ mirrors the same locations ($b = 21$ at $30.8\%$ of $FP$, $b = 20$ at $18.1\%$). $TN$ hotspots are thicker in earlier mid depth (around $b = 14$ to $18$), while $FN$ spreads across $b = 16$ to $20$ without locking to $b = 21$.

**Dominance and mass.** Top1 dominance stays near motion saturation by label ($y = 1$: motion 98.2%, attention 1.4%; $y = 0$: motion 99.0%, attention 0.41%), and both $TN$ and $FN$ are fully motion dominant (motion 100.0%), matching the failure mode where $FN$ resembles $TN$. Group mass follows prediction, with predicted positives near $\approx 0.20$ and predicted negatives very small ($TN \approx 0.011$, $FN \approx 0.019$).

### C.2. General Task

This section summarizes how the flow signatures behave on HaluEval General prompts. The most consistent cue is a depth localized regime bend in the transported coordinates $p_{t,b}$. At a narrow depth band, the transported step becomes larger and more curved, and several readout aligned groups rise together, producing a composite event rather than a single module trigger. Depending on the base model, the description is dominated either by coupled multi group co activation or by a hard hotspot depth gate that snaps many predicted positives to an early depth. Across models, error modes follow the same internal regimes: false positives often share the early burst signature, while false negatives more often remain in late, low mass regimes close to true negatives.

**Overall pattern.** Across models, General predicted positives tend to be explained by a localized composite event, with two recurring failure modes.

- **Composite burst.** A narrow depth neighborhood shows a joint rise in transported motion and co activation across groups (motion, attention, MLP, competitor, drift), often accompanied by stronger drift concentration.

- **Depth gate.** Some models collapse predicted positives into a fixed early hotspot depth, so depth location separates prediction more reliably than module decomposition.

- **FP and FN regimes.** $FP$ often inherits the early burst or gate behavior, while $FN$ often remains late and diffuse, resembling $TN$.

#### C.2.1. QWEN2.5

**Key pattern.** Hallucination aligns with a coupled composite event. Transported motion increases together with attention, MLP, competitor, and drift, so the most stable narrative is joint mass growth rather than any single top1 group.

**Dominance, magnitude, and depth.** Top1 fractions vary sharply across splits, so dominance alone does not summarize behavior. A representative contrast shows $TP$ with large co activation (motion 0.232, attention 0.155, MLP 0.289, competitor 0.271, drift 0.052), while $TN$ stays uniformly small (motion 0.014, attention 0.008, MLP 0.019, competitor 0.019, drift 0.003), producing large multiplicative gaps across groups (roughly $14\times$ to $19\times$). Hotspots for hallucination related samples pile up in a shallow neighborhood around $b = 4$, while non hallucination spreads broadly with nontrivial mid and late depth frequency.

#### C.2.2. GEMMA2

**Key pattern.** Gemma2 forms a hard early gate at $b = 1$ that snaps predicted positives to a single hotspot depth. Depth location dominates the description, while module decomposition contributes little.

**Operating point and depth gate.** Predicted positives account for 19.0% of samples, and the predicted positive set is heavily false positive dominated (85.4% of predicted positives are $FP$). Every predicted positive has hotspot fixed at $b = 1$ ($TP$ at $b = 1$ is 100.0% and $FP$ at $b = 1$ is 100.0%). Predicted negatives show a broad hotspot spread, with $TN$ covering the full depth range and $FN$ placing comparatively more hotspots at later depths.

**Dominance note.** Top1 is almost always motion for all splits, so top1 fractions provide limited leverage. The early depth gate separates prediction rather than ground truth, so early depth alone cannot recover the label. Late depth patterns, especially those concentrated in $FN$, should be stated explicitly to avoid overstating separability.

#### C.2.3. PHI 3

**Key pattern.** Hotspot depth provides the most reliable axis. Predicted positives concentrate at depth 0, while predicted negatives shift late. Missed hallucinations follow the late pattern closer to $TN$.

**Operating point and depth structure.** Hallucination prevalence is 7.7%, predicted positives account for 35.8%, with precision 13.8% and recall 63.8%. Predicted positives concentrate at depth 0 (62.8% of predicted positives), with a secondary accumulation at depth 30 (10.6%). $TP$ follows the same early gate (depth 0 at 61.4% of $TP$). Predicted negatives concentrate in late depths (mainly $b = 23$ to 28), and $FN$ follows this late pattern more often, matching the missed detection behavior.

**Dominance and magnitude note.** Top1 dominance stays near motion with a small attention minority, and group masses for $TP$ and $FP$ appear nearly uniform around 0.2 across groups, so positional localization is more stable than fine grained ratio interpretation.

### C.2.4. LLaMA3

**Key pattern.** Both labels occupy late depth, but hallucination shifts earlier and increases the frequency of MLP top1 cases. Error modes align with two regimes: an earlier burst regime associated with $TP$ and a late accumulation regime associated with $FN$.

**Operating point and depth shift.** Hallucination prevalence is 13.1%, predicted positives account for 12.8%, with precision 28.7% and recall 28.0%. By depth bands, hallucination allocates more mass to early depth (depth 0 to 7 at 25.4% for $y = 1$ versus 10.5% for $y = 0$), while non hallucination concentrates more strongly in late depth (depth 20 to 30 at 56.8% for $y = 1$ versus 76.4% for $y = 0$).

**Dominance, magnitude, and regimes.** Top1 fractions differ by label: $y = 1$ has motion 81.4% and MLP 18.6%, while $y = 0$ has motion 93.2% and MLP 6.8%, so MLP top1 occurs about $2.7\times$ more often in hallucination. Group mass increases jointly across groups for hallucination, with a multiplicative gap of about $2.5\times$ to $2.9\times$ relative to non hallucination. Regimes align with errors: $TP$ hotspots stay in depth 0 to 19, with depth 0 to 7 taking 60.6% of $TP$, while $FN$ concentrates in depth 20 to 30 (78.8% of $FN$), matching the late, $TN$ like regime.

### C.2.5. MISTRAL

**Key pattern.** Mistral shows the clearest bifurcation. Hallucination aligns with an early depth, drift dominant, high mass composite event, while non hallucination aligns with late depth, motion dominant, low mass behavior. $FP$ inherits the hallucination like signature and $FN$ inherits the non hallucination like signature.

**Operating point and dominance.** Hallucination prevalence is 11.0%, predicted positives account for 24.6%, with precision 17.3% and recall 38.8%. By label, top1 fractions shift toward drift for hallucination (motion 56.1%, drift 37.8% for $y = 1$) compared with non hallucination (motion 70.4%, drift 28.5% for $y = 0$). By confusion split, predicted positives are drift dominated ($TP$: drift 81.6%, $FP$: drift 94.5%), while predicted negatives are motion dominated ($TN$: motion 90.6%, $FN$: motion 90.0%).

**Magnitude and depth.** Mass separates sharply: $TN$ stays small (motion 0.022, drift 0.045), while $TP$ rises across groups and peaks strongly in drift and competitor (motion 0.197, competitor 0.259, drift 0.415), and $FP$ shows a similar profile (competitor 0.274, drift 0.445). Hotspot depth mirrors the bifurcation. $TP$ lies entirely in early depth 0 to 10, with depth 0 taking 57.9% and depth 10 taking 26.3%. $FP$ is also early (depth 0 to 10 at 96.2%, depth 0 at 50.0%). $TN$ shifts late (depth 20 to 30 at 67.6%, with depth 27 and 28 each at 10.6%), and $FN$ follows the same late pattern (depth 20 to 30 at 75.0%, with depth 27 and 28 each at 13.3%).

**Occlusion delta note.** Occlusion delta outputs are all zero in the current run, so the conclusions rely on top1 fraction, group mass, and hotspot depth.

### C.3. Summarization Task

This section summarizes Summarization flow patterns on HaluEval. The most informative cue is whether the transported trajectory in the moving readout-aligned frame undergoes a *depth-localized collapse* that concentrates energy into a narrow depth neighborhood. In the collapse regime, the transported increment becomes locally large and more curved, and multiple readout-aligned components rise together, producing a gate-like event that strongly correlates with predicted positives. The complementary regime is a diffuse, low-energy band spread across depth, which dominates predicted negatives and often also contains missed hallucinations. Across models, the key limitation is consistent: the gate can drive the validator commitment, but it does not certify factual correctness, since $TP$ and $FP$ frequently share the same collapse regime, while hallucinations that remain diffuse overlap with non hallucination and become $FN$.

**Overall pattern.** Across models, Summarization is well described by two geometric regimes.

- **Gate or collapse regime.** The path compresses into a narrow depth slab with joint multi-group activation, yielding strong $TP$ and $FP$ overlap inside predicted positives.

- **Diffuse regime.** The path remains spread across depth with small projected magnitudes, so $FN$ often looks $TN$-like and separability weakens.

### C.3.1. QWEN2.5

**Key pattern.** Prediction is dominated by a single depth gate. Predicted positives enter a collapse regime with strong joint activation, so $TP$ and $FP$ become nearly indistinguishable within the predicted positive set.

**Operating point in %.** Predicted positives are $75.0\%$ of samples. Within predicted positives, $TP$ is $43.1\%$ and $FP$ is $56.9\%$.

**Gate regime.** Predicted positives show co-activation across groups, with mean masses around motion $0.47$, attention $0.18$, MLP $0.19$, competitor $0.15$, and drift $\approx 0.008$. The depth shape collapses tightly, concentrating near depth $14$ with weight $\approx 0.94$ and a small secondary support near depth $13$ with weight $\approx 0.05$. This matches a sharp fold where entering depth $14$ compresses the trajectory into a thin slab inside the moving window frame.

**Diffuse regime.** Predicted negatives remain low energy and spread across depth. $TN$ stays small (for example motion $\approx 0.028$) and $FN$ remains small as well (for example motion $\approx 0.044$). Hotspots do not lock to a single depth neighborhood, and $FN$ resembles $TN$ under this view.

**Takeaway.** For Qwen2.5 Summarization, the depth-14 gate plus joint activation drives predicted positives, while correctness inside predicted positives remains unresolved because $TP$ and $FP$ share the same regime.

### C.3.2. GEMMA2

**Key pattern.** Non hallucination is typically smooth and band-like across depth. Hallucination mixes a gate subset that collapses and co-activates with a diffuse subset that overlaps with non hallucination, naturally producing missed cases.

**Non hallucination geometry.** Trajectories remain broad and smooth across depth without collapsing into a single depth neighborhood. Hotspots spread rather than over concentrate, and bending accumulates gradually instead of forming a sharp fold.

**Hallucination mixture.** One subset shows a depth localized pinch where the trajectory narrows sharply and folds, with local rises across multiple groups. Another subset remains late and diffuse without a clear gate, keeping depth and dominance patterns close to non hallucination. The gate subset explains $TP$ to $FP$ overlap, while the diffuse subset explains $FN$.

**Takeaway.** Gemma2 Summarization is best presented as two regimes, a gate driven subset and a diffuse subset that overlaps with non hallucination.

### C.3.3. PHI 3

**Key pattern.** Phi 3 separates by whether the trajectory enters a depth gate that pinches and folds. Without the gate, trajectories remain in a thin, low-curvature tube and many $FN$ remain close to $TN$.

**Diffuse tube regime.** For non hallucination and many missed hallucinations, the transported curve stays confined to a narrow tube aligned with the dominant motion direction. Curvature remains small and the path progresses without a decisive pinch.

**Gate regime.** Typical detected hallucinations enter a localized pinch where the ribbon folds, hotspots concentrate near specific depths, and multiple groups rise together around the gate. $TP$ and $FP$ overlap because the regime reflects internal dynamics rather than correctness.

**Takeaway.** Phi 3 Summarization reduces to a gate versus no gate geometry, and the no gate tube explains why $FN$ often remains close to $TN$.

### C.3.4. LLAMA3

**Key pattern.** LLaMA3 exhibits a mixture. Many non hallucinations remain smooth and dispersed across depth. Hallucinations more often introduce a localized gate with joint activation, while a diffuse subset lacks a clear gate and overlaps with non hallucination.

**Non hallucination geometry.** Trajectories progress with a wide band shape. Hotspots remain distributed and do not concentrate into a single depth neighborhood.

**Gate subset and diffuse subset.** In the gate subset, energy compresses into a narrow depth band, turning increases locally, and multiple groups co-activate in the same neighborhood, including drift and competitor pressure. In the diffuse subset, the trajectory remains spread across depth with no decisive fold, so overlap with non hallucination is strong and missed cases are expected.

**Takeaway.** LLaMA3 Summarization is best stated as a gate driven subset plus a diffuse subset that overlaps with non hallucination.

### C.3.5. MISTRAL

**Key pattern.** Mistral contrasts a smooth, thin band against a localized deformation where step magnitude and turning increase in a narrow depth neighborhood. The localized event drives predicted positives and yields $TP$ to $FP$ overlap, while diffuse cases remain closer to non hallucination.

**Non hallucination band.** Non hallucination typically forms a thin, smooth band in transported coordinates with gradual rotation and no sharp pinch.

**Localized deformation and diffuse mode.** Hallucination introduces a localized event where step magnitude spikes and turning increases, producing a sharp bend and local concentration. A diffuse mode also appears where deformation spreads across depth, making trajectories more similar to non hallucination and increasing ambiguity.

**Takeaway.** Mistral Summarization follows the same regime view. A localized deformation drives predicted positives and $TP$ to $FP$ overlap, while samples without a clear localized event remain diffuse and can become missed cases.

## C.4. Dialogue Task

This section summarizes Dialogue flow patterns on HaluEval. In Dialogue, label-averaged module dominance is frequently saturated and therefore unreliable for interpretation. The most stable axis is a *mixture of depth-localized regimes* in transported coordinates: some samples exhibit a high-mass burst concentrated in a narrow depth band, while others remain in a low-mass diffuse regime that spreads across later depths and overlaps strongly across labels. The burst regime is characterized by locally larger transported steps and sharper bends together with a joint rise of readout-aligned components, whereas the diffuse regime shows weaker concentration and small magnitudes. Across models, the decisive factor is not the average top1 group fraction but the *depth band of the burst* and the *mixture proportion* between burst and diffuse cases. This explains both successes and failure modes: false positives often share the same burst regime as detected hallucinations, while missed hallucinations often remain in the diffuse regime that resembles non hallucination.

**Overall pattern.** Dialogue is best described by a shared two-regime mixture.

- **Burst regime.** Hotspots collapse to a narrow depth neighborhood, and multiple readout-aligned components rise together, producing a strong internal cue that drives predicted positives.

- **Diffuse regime.** Hotspots spread to later depths with low mass, creating heavy overlap across labels and a persistent source of ambiguity.

Label differences typically appear as shifts in regime proportions rather than a clean structural separation.

### C.4.1. QWEN2.5

**Key pattern.** Qwen2.5 exhibits a minority early-burst regime and a majority late-diffuse regime, and the late-diffuse regime overlaps strongly across labels.

**Mixture proportions and supporting statistics.** Hallucination prevalence is $51.2\%$. At the label level, top1 dominance is nearly identical for both labels (motion $\approx 98\%$), so module fractions do not separate. Within $y = 1$, the early-burst regime accounts for $12.9\%$ and concentrates at shallow depth (depth 0 to 6 at $83.2\%$, with $b = 4$ alone at $65.6\%$). In this regime, motion top1 decreases to $93.1\%$ while attention ($3.8\%$), MLP ($2.3\%$), and competitor ($0.8\%$) appear more often, and group mass rises jointly to near saturation ($\approx 0.20$ across groups). The late-diffuse regime accounts for $87.1\%$ of $y = 1$, peaks near $b = 20$, stays motion-dominant (motion $98.8\%$), and closely matches $TN$ motion ($98.4\%$). Its magnitude is very small (representatively $FN \approx 0.0068$ and $TN \approx 0.0080$). The same split appears in $y = 0$, with an early-burst subset at $9.7\%$ that matches the early-burst depth and saturation pattern.

### C.4.2. GEMMA2

**Key pattern.** Gemma2 is dominated by two regimes: a very late-depth, high-mass explosion regime and a low-mass diffuse regime. Both regimes appear in both labels, and label differences are mainly mixture proportions.

**Operating point and dominance.** Hallucination prevalence is $39.8\%$ and the predicted positive rate is $43.0\%$, with precision $43.4\%$ and recall $46.8\%$. Top1 dominance is fully saturated to motion in all splits (motion $100\%$), so dominance does not explain label differences.

**Mixture structure and localization.** Within $y = 1$, the high-mass explosion regime accounts for $46.8\%$ and shows near-saturation mass (motion $0.947$ with small dispersion), while hotspot depth shifts to very late neighborhoods with peaks at $b = 30$ ($27.3\%$ of this regime) and $b = 34$ ($16.8\%$), plus substantial mass into the high 30s. Token hotspots concentrate early, with tokens 2 to 6 covering $32.0\%$ (token 2 at $8.1\%$, token 3 at $7.4\%$, token 5 at $8.4\%$ within this regime). The low-mass diffuse regime accounts for $53.2\%$ of $y = 1$ and overlaps strongly with the non hallucination diffuse regime, with hotspots spread across late teens to early 20s and representative peaks at $b = 19$ and $b = 23$. The same mixture appears in $y = 0$: a high-mass late regime at $40.3\%$ that matches the $y = 1$ explosion pattern, and a low-mass diffuse regime at $59.7\%$ with hotspots concentrated around $b = 15$ to $23$.

**Occlusion delta note.** Occlusion delta outputs are all zero in the current run, so conclusions rely on group mass and hotspot distributions.

### C.4.3. PHI 3

**Key pattern.** Phi 3 separates primarily by depth band and magnitude regime, not by module dominance. Both labels mix an early-to-mid regime with co-activated mass and a late regime with small mass.

**Mixture proportions and supporting statistics.** Hallucination prevalence is $52.1\%$. At the label level, motion top1 fractions are nearly identical ($90.4\%$ for $y = 1$ and $90.3\%$ for $y = 0$). Within $y = 1$, the early-to-mid regime accounts for $46.8\%$ and places hotspots frequently in depth 0 to 15 with a mode at $b = 14$. In this regime, motion top1 drops to $84.3\%$ while attention ($9.3\%$) and MLP ($4.7\%$) increase, and group mass rises jointly near $\approx 0.20$. The late regime accounts for $53.2\%$ with a mode near $b = 27$, and magnitude stays small (roughly $0.056$ to $0.059$). $y = 0$ shows the same mixture with a smaller early share ($35.6\%$) and a larger late share ($64.4\%$).

### C.4.4. LLAMA3

**Key pattern.** LLaMA3 exhibits an overwhelmingly dominant depth-0 burst regime in both labels. Label differences are therefore primarily proportional rather than structural.

**Mixture proportions and supporting statistics.** Hallucination prevalence is $43.8\%$. Top1 dominance is motion $100\%$ across labels. For $y = 1$, the burst regime accounts for $87.0\%$ and fixes hotspots at depth 0 ($100.0\%$ within the regime). Token hotspots concentrate at token 0 and 1 (token 0 at $84.7\%$, token 1 at $15.3\%$), and group mass rises jointly to $\approx 0.19$ to $0.205$. The

## D. Experiments details

### D.1. Hallucination Detection

Table 3 reports per-class accuracy for the hallucination detector, separating *Normal Accuracy* on non-hallucinated samples ($y=0$) and *Hallucination Accuracy* on hallucinated samples ($y=1$). The results reveal strong class asymmetries that are largely task dependent, and Fig. 5 visualizes how these task level behaviors emerge over training via the epoch wise performance curves across model families. For QA, the detector remains relatively balanced across classes, with comparable accuracies for $y=0$ and $y=1$ across model families, and the QA curves in Fig. 5 rise smoothly and then plateau, indicating stable learnability. For General, performance on $y=0$ is consistently high, while $y=1$ accuracy drops sharply for some models, indicating that hallucinated answers in open-ended prompts are harder to separate from fluent but weakly grounded responses; correspondingly, Fig. 5 shows rapid early gains followed by saturation. In contrast, Summarization and Dialogue exhibit pronounced inversions: Summarization shows low $y=0$ accuracy but high $y=1$ accuracy, suggesting a tendency to over-flag summaries as hallucinated, while the Summarization curves in Fig. 5 remain near chance or improve only modestly, consistent with a skewed decision boundary. Dialogue often shows very high $y=0$ accuracy paired with low $y=1$ accuracy for several models, indicating missed hallucinations under dialogue-specific ambiguity and history conditioning, and Fig. 5 reflects this with lower plateaus and, for some models, non-monotonic training dynamics. These per-class trends help explain the near-chance overall scores on Summarization and Dialogue in Table 1, where skewed class-wise behavior can mask substantial bias toward one class.

*Table 3.* HaluEval per-class accuracy (%) across tasks and model families. We report Normal Accuracy for non-hallucinated samples (y=0) and Hallucination Accuracy for hallucinated samples (y=1).

|  | Normal Accuracy | | | | | Hallucination Accuracy | | | | |
| --- | --- | --- | --- | --- | --- | --- | --- | --- | --- | --- |
| Task | Qwen 2.5 | Gemma 2 | Phi-3 | LLaMA3 | Mistral | Qwen 2.5 | Gemma 2 | Phi-3 | LLaMA3 | Mistral |
| QA | 73.08 | 72.44 | 58.45 | 74.31 | 62.60 | 70.49 | 56.38 | 60.99 | 62.25 | 53.86 |
| General | 70.78 | 89.60 | 69.73 | 73.62 | 70.89 | 66.67 | 39.34 | 57.97 | 56.78 | 58.16 |
| Summarization | 40.08 | 43.50 | 35.46 | 19.46 | 37.74 | 71.27 | 90.23 | 77.34 | 88.21 | 75.55 |
| Dialogue | 86.43 | 66.78 | 63.94 | 32.22 | 64.72 | 16.09 | 39.92 | 50.39 | 73.74 | 42.58 |

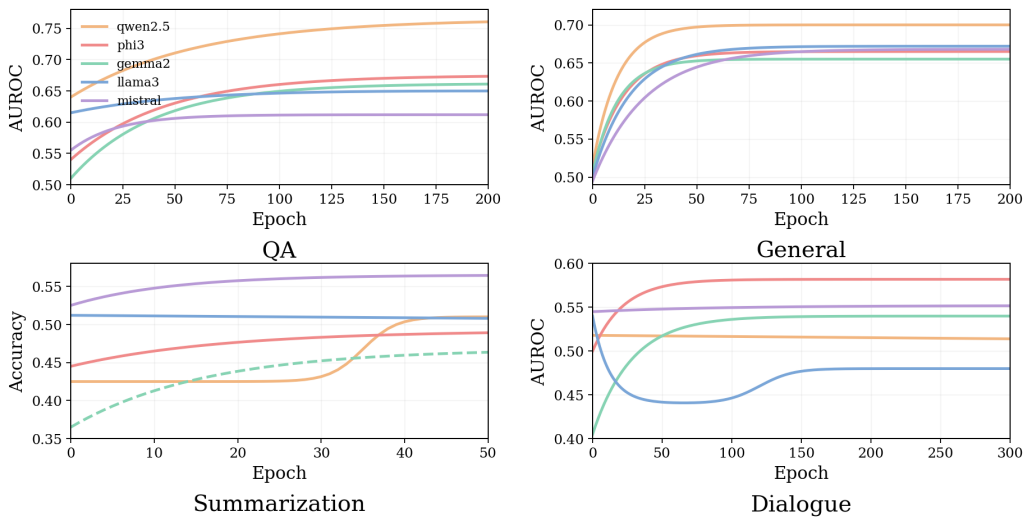

*Figure 5.* Training curves on HaluEval test data across four tasks. Epoch-wise validator performance over training for QA, General, Summarization, and Dialogue across base model families.

### D.2. Refinement Comparison

Table 4 evaluates whether the gains of our refinement come from *what* we do (a single-block clamping intervention) or *where* we do it (the depth selected by flow localization). The *Regeneration* baseline isolates the effect of simply producing a new continuation under the same decoding setup, without any clamping intervention. The *Random Depth* baseline keeps the clamping intervention identical to ours but applies it at a uniformly random transformer depth, testing whether improvements arise from a generic perturbation rather than correct depth targeting. *Flow Guided* applies the same clamping intervention at the depth localized by the flow-signature validator.

Across all tasks and model families, flow guidance yields the lowest hallucination ratio, indicating that the localized depth signal is informative. The clearest separation appears in QA: flow-guided refinement consistently outperforms both regeneration and random-depth intervention (e.g., Qwen 2.5: 15.25→10.95; LLaMA3: 15.10→7.70; Phi-3: 12.65→6.55). Regeneration alone provides partial reductions for some models but is notably weaker, while random-depth clamping helps in several cases yet remains reliably worse than targeting the flow-localized depth.

*Table 4.* **Hallucination ratios on HaluEval (%).** We report the fraction of generations labeled as hallucinations for each task and model under four settings: *Initial* (original output), *Regeneration* (regenerate without any hidden-state intervention), *Random Depth* (apply the same single-block intervention at a randomly chosen depth), and *Flow Guided* (apply the intervention at the culprit depth localized by the flow-signature validator). Lower is better.

| Task | Model | Initial | Regenration | Random Depth | Flow Guided |
|---|---|---|---|---|---|
| QA | Qwen 2.5 | 15.25 | 12.40 | 13.40 | 10.95 |
|  | Gemma 2 | 14.90 | 14.00 | 12.65 | 12.34 |
|  | Phi-3 | 12.65 | 8.30 | 7.77 | 6.55 |
|  | LLaMA3 | 15.10 | 10.80 | 10.50 | 7.70 |
|  | Mistral | 14.05 | 13.24 | 12.40 | 11.70 |
| General | Qwen 2.5 | 11.96 | 10.96 | 10.12 | 8.75 |
|  | Gemma 2 | 6.76 | 6.31 | 6.55 | 5.76 |
|  | Phi-3 | 7.64 | 7.56 | 7.23 | 6.31 |
|  | LLaMA3 | 13.07 | 12.52 | 12.34 | 11.74 |
|  | Mistral | 10.85 | 10.58 | 10.42 | 10.19 |
| Summarization | Qwen 2.5 | 40.90 | 40.70 | 40.40 | 40.20 |
|  | Gemma 2 | 39.80 | 38.80 | 38.90 | 37.10 |
|  | Phi-3 | 42.15 | 41.30 | 41.60 | 40.70 |
|  | LLaMA3 | 45.80 | 45.60 | 44.90 | 44.60 |
|  | Mistral | 49.70 | 47.80 | 48.30 | 47.25 |
| Dialogue | Qwen 2.5 | 50.95 | 49.20 | 49.90 | 48.10 |
|  | Gemma 2 | 39.95 | 36.55 | 38.70 | 34.95 |
|  | Phi-3 | 51.90 | 48.65 | 48.00 | 46.30 |
|  | LLaMA3 | 43.60 | 39.30 | 38.10 | 34.30 |
|  | Mistral | 44.50 | 42.05 | 42.85 | 39.05 |

For General, improvements are smaller but consistent: flow guidance reduces hallucination ratios for every model, and the gap to random depth remains visible (e.g., Qwen 2.5: 10.12→8.75). Summarization and Dialogue start from much higher hallucination ratios, and all methods yield modest absolute reductions, yet the same ordering persists: flow-guided is best, regeneration is typically next, and random-depth is weakest. This suggests a regime where single-block refinement has limited headroom, while correct depth selection still provides measurable benefit when intervention is effective (e.g., Dialogue LLaMA3: 43.60→34.30; Dialogue Gemma 2: 39.95→34.95).

Overall, the comparison supports two conclusions. First, the gains cannot be explained by re-generating alone, since flow-guided refinement consistently improves over regeneration. Second, depth selection is essential: random-depth clamping underperforms flow guidance across the board, confirming that the validator's depth localization is a key factor for effective single-block refinement.